\documentclass[journal]{IEEEtran}
\usepackage{amsfonts}
\usepackage[nocompress]{cite}
\usepackage[cmex10]{amsmath}
\usepackage{color}
\usepackage{flushend}
\usepackage{url}
\usepackage{array}

\usepackage[table,xcdraw]{xcolor}
\usepackage{lineno}
\usepackage{flushend}
\usepackage{amsfonts}
\usepackage{mdwmath}
\usepackage{booktabs}
\usepackage{rotating}

\definecolor{rblue}{rgb}{0,0.5,1}
\usepackage[colorlinks]{hyperref}
\hypersetup{linkcolor=red,citecolor=rblue}
\usepackage{subfigure}   %figure bing lie
\usepackage{graphicx}    %figure bing lie
\usepackage{multirow}
\usepackage{makecell}
\usepackage{float}
\usepackage{dsfont}
\usepackage{bbding}

\definecolor{mygreen}{rgb}{0, 0.6, 0}
\definecolor{mypurple}{rgb}{0.75, 0, 0.25}
\definecolor{aliceblue}{rgb}{0.94, 0.97, 1.0}
\definecolor{lightpurple}{rgb}{0.498, 0.459, 0.647}

\newcommand{\improved}[1]{\textcolor{mygreen}{#1}}

\newcommand{\revised}[1]{\textcolor{black}{#1}}
\newcommand{\marked}[1]{\textcolor{black}{#1}}
 
\begin{document}
\title{EchoTrack: Auditory Referring Multi-Object Tracking for Autonomous Driving}
\author{Jiacheng Lin, 
        Jiajun Chen, 
        Kunyu Peng, 
        Xuan He, 
        Zhiyong Li,~\IEEEmembership{Member,~IEEE}, \\
        Rainer Stiefelhagen,~\IEEEmembership{Member,~IEEE},
        and Kailun Yang%
\thanks{This work was supported in part by the National Natural Science Foundation of China under Grant U21A20518 and Grant U23A20341; and in part by Hangzhou SurImage Technology Company Ltd. \textit{(Jiacheng Lin, Jiajun Chen, and Kunyu Peng contributed equally to this work.)} \textit{(Corresponding authors: Zhiyong Li and Kailun Yang.)}
}
\thanks{J. Lin, X. He, and Z. Li are with the College of Computer Science and Electronic Engineering, Hunan University, Changsha 410082, China. (e-mail: jcheng\_lin@hnu.edu.cn; mikasa@hnu.edu.cn; zhiyong.li@hnu.edu.cn).}
\thanks{J. Chen, Z. Li, and K. Yang are with the School of Robotics, Hunan University, Changsha 410012, China. (e-mail: chenjiajun@hnu.edu.cn; zhiyong.li@hnu.edu; kailun.yang@hnu.edu.cn).}
\thanks{J. Chen, Z. Li, and K. Yang are also with the National Engineering Research Center of Robot Visual Perception and Control Technology, Hunan University, Changsha 410082, China.  (e-mail: chenjiajun@hnu.edu.cn; zhiyong.li@hnu.edu; kailun.yang@hnu.edu.cn).}
\thanks{K. Peng and R. Stiefelhagen are with the Institute for Robotics and Anthropomatics, Karlsruhe Institute of Technology, 76131 Karlsruhe, Germany. (e-mail: kunyu.peng@kit.edu; rainer.stiefelhagen@kit.edu).}
}

% The paper headers
\markboth{IEEE Transactions on Intelligent Transportation Systems, August~2024}%
{Lin \MakeLowercase{\textit{et al.}}: EchoTrack}

\maketitle

\begin{abstract}
This paper introduces the task of Auditory Referring Multi-Object Tracking (AR-MOT), which dynamically tracks specific objects in a video sequence based on audio expressions and appears as a challenging problem in autonomous driving. Due to the lack of semantic modeling capacity in audio and video, existing works have mainly focused on text-based multi-object tracking, which often comes at the cost of tracking quality, interaction efficiency, and even the safety of assistance systems, limiting the application of such methods in autonomous driving. In this paper, we delve into the problem of AR-MOT from the perspective of audio-video fusion and audio-video tracking. We put forward EchoTrack, an end-to-end AR-MOT framework with dual-stream vision transformers. The dual streams are intertwined with our Bidirectional Frequency-domain Cross-attention Fusion Module (Bi-FCFM), which bidirectionally fuses audio and video features from both frequency- and spatiotemporal domains. Moreover, we propose the Audio-visual Contrastive Tracking Learning (ACTL) regime to extract homogeneous semantic features between expressions and visual objects by learning homogeneous features between different audio and video objects effectively. Aside from the architectural design, we establish the first set of large-scale AR-MOT benchmarks, including Echo-KITTI, Echo-KITTI+, and Echo-BDD. Extensive experiments on the established benchmarks demonstrate the effectiveness of the proposed EchoTrack and its components. The source code and datasets are available at \url{https://github.com/lab206/EchoTrack}.
\end{abstract}

\IEEEpeerreviewmaketitle

\section{Introduction}
\label{sec:introduction}
\IEEEPARstart{R}{eferring} scene understanding~\cite{wu2023language, pan2022wnet,tang2021multi,ding2024text} has garnered significant attention within the vision community, primarily due to its potential applications in domains, \textit{i.e.}, autonomous driving~\cite{wu2023referring,nguyen2023type} and image editing~\cite{zhou2022audio,mao2023multimodal}.
Referring Multi-Object Tracking (RMOT)~\cite{du2023ikun, wu2023referring, nguyen2023type, zheng2023towards}, as one of the dominant tasks in \revised{referring scene understanding}, grasps large attention from the community as its superior assistance in diverse application scenarios, \textit{i.e.}, surveillance~\cite{nguyen2023type,elhoseny2020multi}, intelligent  vehicles~\cite{wu2023language,wu2023referring}, and robotics~\cite{zhao2023transformer}, offer tangible benefits for Intelligent Transportation Systems (ITS)~\cite{feng2021cityflow}.

With paramount significance in numerous domains, RMOT plays a pivotal role in enhancing situational awareness, decision-making, and safety. 
%<*response-r13-1>
Currently, most of the existing works are conducted based on text-based inference, \revised{\textit{i.e.}, TransVLT~\cite{zhao2023transformer}, TransRMOT~\cite{wu2023referring}, and GMOT~\cite{nguyen2023type}}.
However, when considering the convenience of interaction in \revised{autonomous driver assistance systems~\cite{wu2023language}} and the accessibility of helping people with visual impairments~\cite{liu2021hida}, text references show clear limitations, resulting in unsatisfactory interaction efficiency and assistance effectiveness. For \revised{autonomous driver assistance systems}, typing text information into the tracking system while driving would downgrade the system's reliability and be detrimental to the driver's safety.
%</response-r13-1>

\begin{figure}[t!]
\centering
\includegraphics[width=1\linewidth]{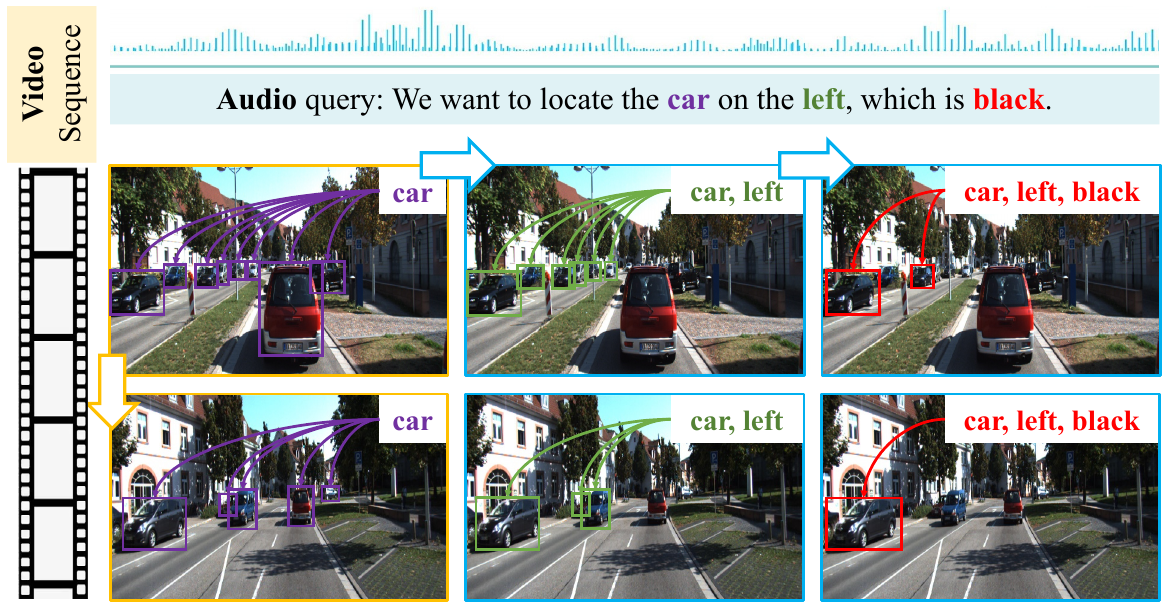}
\caption{Overview of the introduced Auditory Referring Multi-Object Tracking (AR-MOT) task. The audio reference and the video are both fed into the model. The model is expected to track objects that are mentioned in the audio input step by step. Two samples from our Echo-KITTI+ dataset are provided to illustrate the workflow of the challenging AR-MOT.}
\label{fig:teaser}
\end{figure}

To tackle the above issues in text-based referring scene understanding methods, we propose an Auditory Referring Multi-Object Tracking (AR-MOT) task, as illustrated in Fig.~\ref{fig:teaser}. AR-MOT aims to dynamically locate the relevant visual objects from the video based on the audio input semantics, which effectively ensures the quality of the interactions as well as efficiency.
To delve into this challenging task, we first establish a set of large-scale AR-MOT benchmarks based on the KITTI~\cite{geiger2012we,wu2023language} and BDD100K~\cite{yu2020bdd100k} datasets. 
The AR-MOT benchmarks, structured into three primary sub-datasets, namely Echo-KITTI, Echo-KITTI+, and Echo-BDD, contain a total of $86$ videos and $23,305$ frames, along with $6,220$ audio and text expressions, respectively. Within each video, there are more than an average of $80$ objects, which provide a variety of distributions of audio length, video difficulty, complex traffic scenarios, and severe challenges in locating the dynamic objects in unconstrained surroundings.

\begin{figure}[t!]
\centering

\includegraphics[width=1\linewidth]{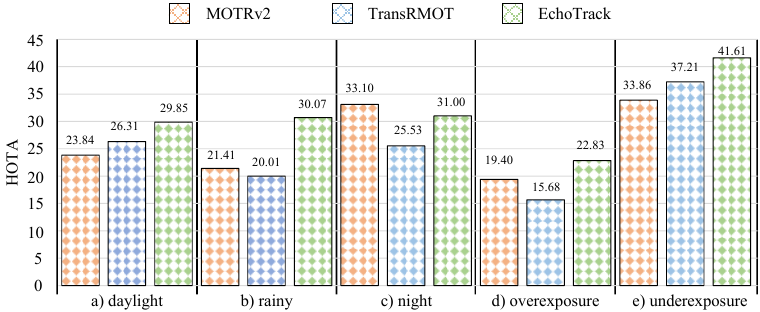}
%<*response-r14>
\caption{\revised{Comparison of tracking performance in HOTA~\cite{luiten2021hota} on the established Echo-BDD dataset with different conditions. MOTRv2~\cite{zhang2023motrv2} is a representative MOT method, and TransRMOT~\cite{wu2023referring} is a RMOT method. Both of them use the HuBERT-Base~\cite{hsu2021hubert} to encode the audio. The proposed EchoTrack consistently outperforms other methods across five conditions.}}
%</response-r14>
\label{fig:fig1-2}
\end{figure}

Aside from constructing a fresh set of AR-MOT benchmarks, training such models is not straightforward, as associating audio cues adequately with multiple dynamic visual objects and avoiding the attenuation of referring features due to long-range propagation are challenging.
Our design is motivated by the observations in Fig.~\ref{fig:fig1-2}, which reveal a notable performance degradation by directly using the methods~\cite{wu2023referring,nguyen2023type} developed specifically for RMOT to tackle the novel AR-MOT task.
This decline in performance can be attributed to the substantial modality disparity existing between text-based and audio-based features. 
Specifically, in addition to expressing complexity, audio is heavily influenced by factors like timbre, speed, and noise, whereas text is not.
It is evident that existing methods~\cite{pan2022wnet, chen2023epcformer, senocak2023sound} derived from text-based \revised{referring scene understanding} methods are ill-suited for audio-based ones due to the inherent differences in the modalities involved.
Consequently, a compelling necessity arises for the development of specialized approaches tailored explicitly to addressing the unique challenges posed by AR-MOT tasks.

Furthermore, to seek an effective way of handling AR-MOT tasks, we make use of the fusion between the spatiotemporal and frequency domains and propose a novel Bidirectional Frequency-domain Cross-Attention Fusion Module (Bi-FCFM) to facilitate transformer-based information aggregation, considering that frequency information contains essential domain cues of the audio data, which provide valuable references. We propose Audio-visual Contrastive Tracking Learning (ACTL), which explores the potential homogeneous semantic information between audio expression and visual objects to construct contrast learning by introducing audio-referring features before output. A homogeneous semantic alignment representation space can be built through the aforementioned procedure. Extensive experimental results on six datasets show that the proposed EchoTrack solution consistently exhibits state-of-the-art tracking performance in both the established AR-MOT and RMOT tasks.

%<*response-r13-2>
At a glance, we deliver the following contributions:
\begin{itemize}
    \item We propose Auditory Referring Multi-Object Tracking (AR-MOT), a referring scene understanding task aiming to dynamically localize relevant visual objects in the video based on the audio expression. 
    \item We establish AR-MOT benchmarks, containing three sub-datasets, Echo-KITTI, Echo-KITTI+, and Echo-BDD, to offer rich referring visual attributes and audio expressions. To the best of our knowledge, this is the first work that delivers AR-MOT benchmarks.
    \item \revised{We introduce EchoTrack, an end-to-end AR-MOT solution incorporating a bidirectional frequency-domain cross-attention fusion module and an audio-visual contrastive tracking learning regime.} Our proposed EchoTrack approach consistently elevates state-of-the-art tracking performance.
\end{itemize}
%</response-r13-2>
    
%<*response-r110-1>
\revised{The remainder of the paper is structured as follows: Sec.~\ref{sec:related_work} provides a brief overview of related work. The proposed methods and benchmarks are presented in Sec.~\ref{method} and Sec.~\ref{sec:datasets}, respectively. We present experimental results in Sec.~\ref{sec:experiment}. Sec.~\ref{sec:conclusion} summarizes the findings presented in this work.}
%</response-r110-1>

\section{Related Work}
\label{sec:related_work}

\subsection{Multi-Object Tracking}
Multi-Object Tracking (MOT)~\cite{yan2023bridging,gao2023memotr,zhang2023motrv2,zhang2023attentiontrack,li2020learning,cao2022multi} is a fundamental problem in computer vision~\cite{liu2023lightweight,li2023human}, which is considered more challenging and applicable in real-world scenarios, \textit{i.e.}, autonomous vehicles~\cite{wu2023language,wu2023referring} and robotics~\cite{zhao2023transformer}, compared to single-object tracking~\cite{sun2020transtrack,meinhardt2022trackformer}. 
MOT can be addressed via diverse techniques. 
For example, spatiotemporal memory~\cite{cai2022memot, mukhtar2023stmmot, xu2019spatial, you2021multi, miah2023multi} is widely leveraged for better MOT reasoning. 
Yu~\textit{et al.}~\cite{yu2022towards} propose a multi-view trajectory contrastive learning for MOT. 
Shuai~\textit{et al.}~\cite{shuai2021siammot}, Gao~\textit{et al.}~\cite{gao2022multi}, and Ma~\textit{et al.}~\cite{ma2021deep} leverage a siamese network to learn discriminative features for MOT. 
Tracklet association is used by~\cite{wang2021track,peng2020tpm,saleh2021probabilistic,bae2014robust, wang2022split,chen2019aggregate, zhang2022bytetrack}. 
Further, graph neural network~\cite{wang2021joint, weng2020gnn3dmot, papakis2020gcnnmatch,li2020graph} is frequently explored for MOT due to its surpassing relation reasoning capability towards multiple components in a scene. 
Motion models~\cite{kesa2021joint, sun2019deep, qin2021joint, zou2022compensation} are used to compute dissimilarity scores according to the object movements. 
Vision transformers~\cite{sun2020transtrack,chen2022patchtrack,yu2022relationtrack,blatter2023efficient,zhou2022global, zhou2022audio,zeng2022motr} play an important role recently in MOT due to their superior long-term context reasoning capacity.
Zeng~\textit{et al.}~\cite{zeng2022motr} propose MOTR to achieve end-to-end MOT with transformers.
Zhang~\textit{et al.}~\cite{zhang2023motrv2} enhance MOTR by using an auxiliary object detector.
Different from existing works, we introduce a challenging AR-MOT task to elevate tracking capacity in unconstrained scenarios.

\subsection{Referring Scene Understanding}
Referring scene understanding encompasses text- and audio-based referring domains, with the former garnering significant research attention.
However, a notable gap exists in the literature on MOT with audio-based references.
Addressing this gap is imperative, emphasizing the need for focused research efforts in this area.

\subsubsection{Text-based Referring Scene Understanding} 
In the realm of text-based referring scene understanding, numerous works have made substantial contributions. Text-based image segmentation~\cite{liu2023polyformer}, video object segmentation~\cite{miao2023spectrum}, expression comprehension~\cite{luo2020multi, jiang2023video_referring_expression_comprehension}, vision language tracking~\cite{zhao2023transformer,du2023ikun} and grounding~\cite{tang2021multi}, text-image re-identification~\cite{ding2024text}, referring scene comprehension~\cite{yan2023universal}, and object tracking~\cite{wu2023referring,nguyen2023type,he2024visual,feng2021cityflow} are well explored by the community in the past.
Wu~\textit{et al.}~\cite{wu2023referring} for the first time propose an RMOT task with an online cross-modal tracker.
Nguyen~\textit{et al.}~\cite{nguyen2023type} propose the GroOT by using text prompts.
%<*response-r3-1>
\revised{He~\textit{et al.}~\cite{he2024visual} introduced DeepRMOT, a cross-modality representation-based approach aimed at mitigating tracking errors due to vision-language mismatches. 
Du~\textit{et al.}~\cite{du2023ikun} developed a knowledge unification network for aligning trajectories with text references in a non-end-to-end manner, which enhances accuracy when integrated with off-the-shelf tracking systems.}
%</response-r3-1>
However, text reference, requiring the user to type in the information, is clearly not convenient in real-world driving scenarios.

\subsubsection{Audio-based Referring Scene Understanding} 
While auditory referring scene understanding has been studied in semantic segmentation~\cite{pan2022wnet, zhou2022audio, mao2023multimodal}, unified referring scene comprehension~\cite{zou2023segment, chen2023epcformer}, and sound source localization tasks~\cite{senocak2023sound,qian2020multiple}, audio-referring MOT is scarcely addressed in the state-of-the-art.
The integration of audio-based references for tracking, in contrast to text-based approaches, presents several distinct advantages.
Audio references offer more natural and contextually rich descriptions, enhancing the intuitiveness of tracking systems.
The adoption of audio references advances accessibility, benefiting individuals with visual impairments by providing a more effective means of interacting with and comprehending their surrounding environments.
In this work, we look into the under-explored AR-MOT, build up the first set of benchmarks, and put forward an end-to-end EchoTrack model to address the new challenge.

\section{Methodology}
\label{method}
\subsection{Overview}
\begin{figure*}
\centering
\includegraphics[width=0.9\linewidth]{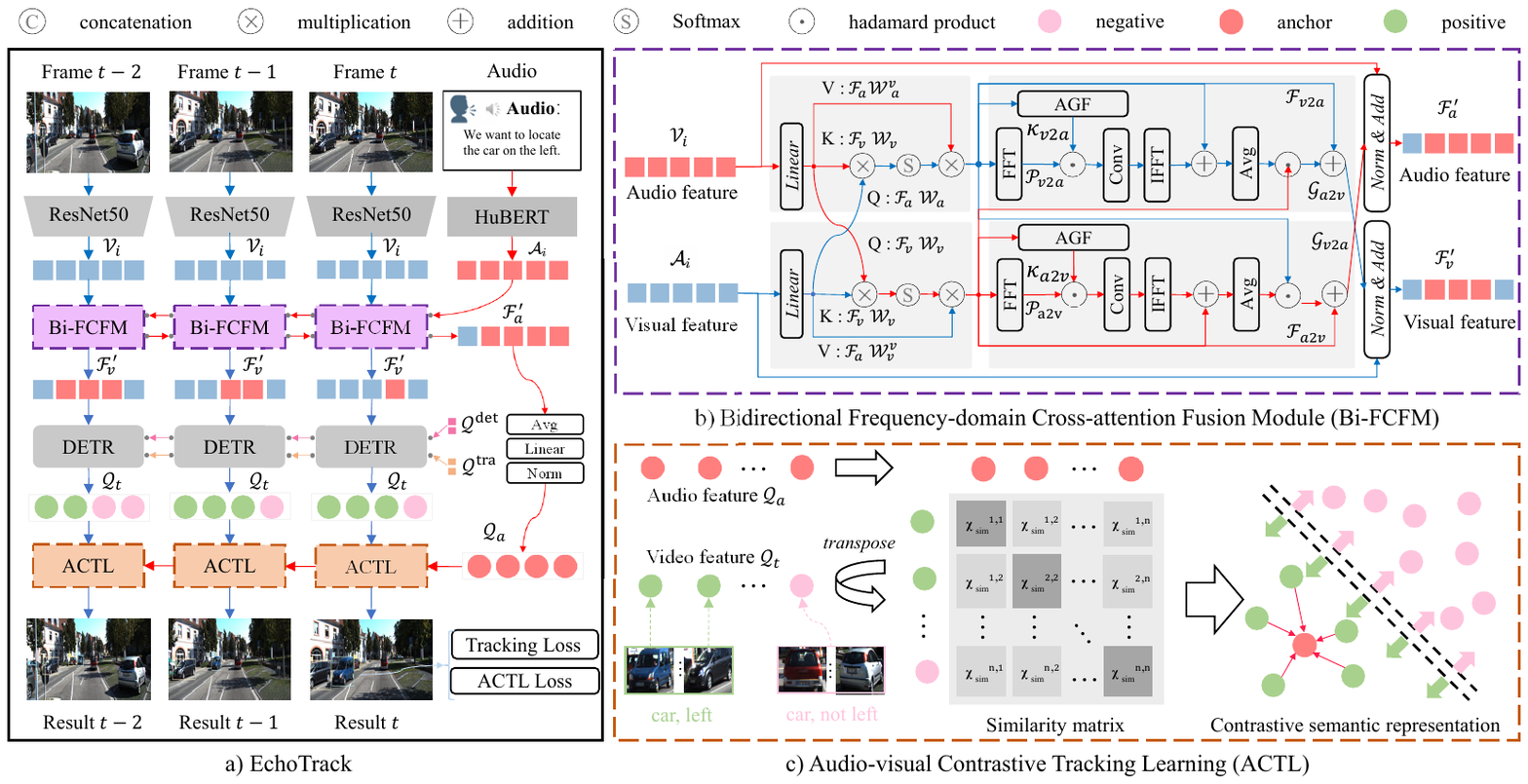}

%<*response-r15>
\caption{\revised{Overview of the proposed EchoTrack.
In a), EchoTrack comprises five primary components: from top to bottom, audio-video encoding, audio-visual feature fusion, audio-visual feature decoding, audio-visual tracking, and matching and loss optimization. In b), AGF stands for the adaptive gaussian filter, and Avg denotes the global average pooling operation.}}
%</response-r15>

\label{fig:framework}
\end{figure*}

To achieve the purpose of tracking relevant objects within a video based on audio expression, we construct an AR-MOT framework, namely EchoTrack, as illustrated in Fig.~\ref{fig:framework}.

\textbf{Audio-video Encoding.}
The encoding operates on a set of video sequences, denoted as $\mathcal{V} = \{\mathcal{V}_i\}_{i=1}^n$, where $\mathcal{V}_i$ represents the visual features encoded using ResNet50~\cite{ResNet} for the $i$-th frame and $n$ indicates the total number of frames. 
Here, each $\mathcal{V}$ is associated with a corresponding audio expression $\mathcal{A} = \{ \mathcal{A}_i \}_{i=1}^m$, with $\mathcal{A}_i$ indicating the feature of the $i$-th audio encoded using frozen HuBERT-Base~\cite{hsu2021hubert} \revised{and $m$ indicates the total number of the audio}.

\textbf{Audio-visual Feature Fusion.}
Existing methods~\cite{pan2022wnet, wu2023language, nguyen2023type, zheng2023towards} have primarily focused on the spatiotemporal domain fusion of audio and video features.
While computationally efficient, these approaches frequently compromise a substantial number of frequency-domain features in both audio and video, impeding the tracking of objects.
Recognizing this, we design the Bidirectional Frequency-domain Cross-Attention Fusion Module (Bi-FCFM) to facilitate transformer-based information aggregation, considering that frequency information contains essential domain cues of the audio data, which provide valuable references.
As shown in Fig.~\ref{fig:framework} b), the proposed Bi-FCFM effectively boosts feature fusion while emphasizing unique frequency-domain traits in audio-visual fusion features, which enables the creation of a shared decision space considering the features of different modalities, \textit{i.e.}, video and audio. In the aforementioned shared decision space, all the features contribute to locating key features relevant to tracking objects in their respective modal features.

\textbf{Audio-visual Feature Decoding.}
%<*response-r1-6>
Similar to~\cite{meinhardt2022trackformer,wu2023referring,zeng2022motr}, we employ deformable DEtection TRansformer (DETR)~\cite{DeformableDETR} with frame propagation mechanism~\cite{zeng2022motr} for audio-visual fusion feature decoding.
To preserve inter-frame object information, we incorporate a frame propagation mechanism based on self-attention, similar to the approach presented in~\cite{zeng2022motr}. This mechanism extends the tracking query from the current frame to the subsequent frame, mitigating the risk of object loss across different frames and facilitating the modeling of object-tracking trajectories.
%</response-r1-6>

\textbf{Audio-visual Tracking.}
The performance of \revised{referring scene understanding} is significantly impacted by the loss of referring features resulting from the long-range propagation of features~\cite{pan2022wnet, wu2023language, zheng2023towards}.
Prior studies have tackled this issue through attention mechanisms~\cite{pan2022wnet, wu2023referring} and employing referring queries~\cite{chen2023epcformer, liu2023polyformer}. However, these methods might be sub-optimal due to ineffective utilization of frequency domain cues and consequent referring feature loss.
To address these limitations, we present an innovative regime named Audio-visual Contrastive Tracking Learning (ACTL). ACTL explicitly promotes interactions between audio and corresponding visual trajectory features.

\textbf{Matching and Loss Optimization.} The EchoTrack includes tracking ($\mathcal{L}_{track}$) and ACTL ($\mathcal{L}_{act}$) losses. Among them, $\mathcal{L}_{track}$ encompasses classification ($\mathcal{L}_{cls}$) and bounding box regression ($\mathcal{L}_{box}$) losses. To be specific, we adopt focal loss to measure $\mathcal{L}_{cls}$ and $\mathcal{L}_{act}$, while $\mathcal{L}_{box}$ consists of the $L1$ ($\mathcal{L}_{l_{1}}$) and IoU ($\mathcal{L}_{iou}$) losses fellow~\cite{wu2023referring,zeng2022motr} and ultimately selects the optimal object through bipartite matching.

\subsection{Bidirectional Frequency-domain Fusion}
As shown in Fig.~\ref{fig:framework} b), the proposed Bi-FCFM consists of three parts: audio-visual spatiotemporal-domain feature fusion, audio-visual frequency-domain representation, and audio-visual frequency-domain feature fusion.

\textbf{Audio-visual Spatiotemporal-domain Feature Fusion.}
We initiate the process by utilizing audio features as an index and performing cross-attention computations with visual features for transformer-based trackers~\cite{wu2023referring,yan2023universal,pan2022wnet,miao2023spectrum}.
Given $\mathcal{V}_i$ and $\mathcal{A}_i$, we project them using linear functions to obtain visual features $\mathcal{F}_v$ and audio features $\mathcal{F}_a$.
Subsequently, we apply the Bi-XAtt mechanism~\cite{yan2023universal} to obtain audio-guided visual features $\mathcal{F}_{a2v}$ as shown in Eq.~\ref{eq:1}.
\begin{equation}
\label{eq:1}
  \mathcal{F}_{v2a} = \text{Softmax}\left(\frac{\mathcal{F}_v\mathcal{W}_v(\mathcal{F}_a \mathcal{W}_a)^\mathrm{T}}{\sqrt{d_i}}\right)(\mathcal{F}_a \mathcal{W}_a^v)^\mathrm{T}.
\end{equation}
Similarly, the image-guided audio features $\mathcal{F}_{a2v} = \text{Softmax}\left(\frac{\mathcal{F}_a \mathcal{W}_a(\mathcal{F}_v \mathcal{W}_v)^\mathrm{T}}{\sqrt{d_j}}\right)(\mathcal{F}_a \mathcal{W}_v^v)^\mathrm{T}$, 
\revised{where $\mathcal{W}_v$, $\mathcal{W}_a$, $\mathcal{W}_v^v$, and $\mathcal{W}_a^v$ are learnable weights, $d_i$ and $d_j$ denote the dimensions of $(\mathcal{F}_a \mathcal{W}_a)$ and $(\mathcal{F}_v \mathcal{W}_v)$, respectively, and $(*)^\mathrm{T}$ is matrix transposition operation.}

\textbf{Audio-visual Frequency-domain Representation.} 
Existing attention-based fusion modules typically prioritize spatiotemporal domain interactions, potentially neglecting the importance of frequency cues, particularly essential low-frequency semantic features~\cite{miao2023spectrum, tatsunami2023fft, chi2020fast}.
While spatiotemporal domain details are crucial, noteworthy cues within the audio domain often reside in the frequency domain realm. This highlights the potential inadequacy of existing attention-based fusion modules for AR-MOT tasks.

To achieve this, we begin by transforming the audio-guided visual features 
$\mathcal{F}_{a2v}$
  and video-guided audio features 
$\mathcal{F}_{v2a}$
from the spatiotemporal into frequency domain spectra using a discrete \revised{Fast Fourier Transform (FFT)} as depicted in Eq.~\ref{eq:3},
\begin{equation}
\label{eq:3}
    \mathcal{P}_{v2a} = \mathcal{F}_{v2a}(k) = \sum_{n = 0}^{N - 1}{f_{v2a}(n)}e^{- j2\pi\frac{kn}{N}},
\end{equation}
where $j$, $k$, $n$, and $N$ denote specific variables, and $f_{v2a}$ signifies the transformed integrated global audio frequency domain spectrum.
Similarly, we derive the transformed global video frequency domain spectrum $\mathcal{P}_{a2v} = \mathcal{F}_{a2v}(k) = \sum_{n = 0}^{N - 1}{f_{a2v}(n)}e^{- j2\pi\frac{kn}{N}}$.

Based on the FFT theorem, $\mathcal{P}_{v2a}$ can be divided into components with low-frequency spectra and high-frequency spectra, denoted as $\mathcal{P}_{v2a} = \left\{ \mathcal{P}_{v2a}^{l},\ \mathcal{P}_{v2a}^{h} \right\}$. 
$\mathcal{P}_{v2a}^{l}$ comprises more abundant chromatic characteristics, whereas $\mathcal{P}_{v2a}^{h}$ encompasses shape contour features.
In this study, we posit that the selection of high- and low-frequency spectra should be adaptable, dynamically adjusting the optimal choice based on individual samples in cross-attention.
To efficiently acquire sample-adaptive filtering kernels, we initially transform the spatiotemporal domain features into the frequency domain decision space using an adaptive non-linear mapping function. By capitalizing on spatial correlations, we dynamically formulate filtering coefficients $\epsilon = \{\epsilon_{v2a},\epsilon_{a2v\ }\}$ and generate adaptive Gaussian filter kernels, as demonstrated in Eq.~\ref{eq:5},
%<*response-r22-1>
\begin{equation}
\label{eq:5}
\mathcal{K}_{v2a} = \frac{\epsilon_{v2a}}{\sigma\sqrt{2\pi}}e^{- \frac{1}{2}\left( \frac{x - \mu}{\sigma} \right)^{2}}.
\end{equation} 
\revised{where $x$ is the random variable, which obeys a normal distribution with mathematical expectation $\mu$ and variance $\sigma$. Similarly, we obtain $\mathcal{K}_{a2v} = \frac{\epsilon_{a2v}}{\sigma\sqrt{2\pi}}e^{- \frac{1}{2}\left( \frac{x - \mu}{\sigma} \right)^{2}}$, where $\epsilon = \{\epsilon_{v2a}=\frac{1}{1 + e^{- \mathcal{M}(\mathcal{F}_{v2a})}}, \epsilon_{a2v}=\frac{1}{1 + e^{- \mathcal{M}(\mathcal{F}_{a2v})}}\}$. $\mathcal{M}$ represents a learnable \revised{multi-layer perceptron.}}
%</response-r22-1>
As per the convolution theorem~\cite{proakis2007digital}, multiplication in the frequency domain is analogous to circular convolution in the spatiotemporal domain, as depicted in Eq.~\ref{eq:6},
\begin{equation}
\label{eq:6}
\begin{aligned}
\mathcal{P}_{v2a}^{'} &= \omega_{v2a}(\mathcal{P}_{v2a} \odot \mathcal{K}_{v2a}), \\
\mathcal{P}_{a2v}^{'} &= \omega_{a2v}(\mathcal{P}_{a2v} \odot \mathcal{K}_{a2v}),
\end{aligned}
\end{equation}
where $\odot$ indicates the Hadamard product. $\omega_{v2a}(\cdot)$ and $\omega_{a2v}(\cdot)$ indicate one-dimensional convolutions, enabling adaptive learning of Gaussian filtering kernels.

To facilitate comprehension of the extracted frequency cues, we utilize \revised{Inverse Fast Fourier Transform (IFFT)} to restore the acquired spectra from the frequency domain back to the spatiotemporal domain, as outlined in Eq.~\ref{eq:7},
\begin{equation}
\label{eq:7}
\mathcal{S}_{v2a} = f_{v2a}(n) = \sum_{n = 0}^{N - 1}\mathcal{F}_{v2a}(k)e^{-j2\pi\frac{kn}{N}}.
\end{equation}
Similarly, we can obtain the $\mathcal{S}_{a2v} = f_{a2v}(n) = \sum_{n = 0}^{N - 1}\mathcal{F}_{a2v}(k)e^{-j2\pi\frac{kn}{N}}$ according to Eq.~\ref{eq:7}. Subsequently, we employ Eq.~\ref{eq:8} to avoid excessive loss of spatiotemporal domain cues, as depicted as follows,
\begin{equation}
\label{eq:8}
\mathcal{F}_{v2a}^{'} = \left[\mathcal{S}_{v2a}, \mathcal{F}_{v2a}\right],
\mathcal{F}_{a2v}^{'} = \left[\mathcal{S}_{a2v}, \mathcal{F}_{a2v}\right],
\end{equation}
where $\left[\cdot, \cdot\right]$ denotes matrix addition.

\textbf{Audio-visual Frequency-domain Feature Fusion.} 
At its core, multimodal interaction entails a cross-correlation filtering process. In AR-MOT tasks, an optimal approach involves the automatic generation of filtering kernels for spatiotemporal domain-enhanced video features via audio frequency domain features.
This filtering accentuates object features that exhibit high correlations with audio features within the video features, as shown in Eq.~\ref{eq:9},
%<*response-r22-2>
\begin{equation}
\label{eq:9}
    \revised{\mathcal{G}_{a2v} = \left [ \frac{1}{L}\sum_{l = 1}^{L} \mathcal{F}_{v2a}^{'}[:, l] \odot \mathcal{F}_{a2v}, \mathcal{F}_{v2a} \right ].}
\end{equation}
\revised{where $l$ is the index number of the length $L$ of the feature. Similarly, we have $\mathcal{G}_{v2a} = \left [ \frac{1}{L}\sum_{l = 1}^{L} \mathcal{F}_{a2v}^{'}[:, l] \odot \mathcal{F}_{v2a}, \mathcal{F}_{a2v} \right ]$.}

Finally, we obtain fused output as shown in Eq.~\ref{eq:10},
\begin{equation}
\label{eq:10}
 \revised{\mathcal{F}_{v}^{'} = \left[\mathcal{G}_{a2v}, \text{Norm}(\mathcal{V}_i)\right],
 \mathcal{F}_{a}^{'} = \left[\mathcal{G}_{v2a}, \text{Norm}(\mathcal{A}_i)\right].}   
\end{equation}
\revised{where Norm indicates the normalization.}
%</response-r22-2>
\subsection{Audio-visual Contrastive Tracking Learning}

In this section, we introduce an innovative approach termed ACTL, encapsulating its core principles illustrated in Fig.~\ref{fig:framework}~c). This method aims to reduce the gap between audio features and referring objects while simultaneously pushing away from non-referring objects. \revised{Specifically, 
we align the dimensions of the averaged pooled audio feature, denoted as ${\ \mathcal{Q}}_{a} = {\text{Avg}\mathcal{(F}'}_{a})$, with those of the tracking trajectory instance query, ${\ \mathcal{Q}}_{t}^{'}$. The Avg indicates the global average pooling operation.} Subsequently, normalization is applied to $\mathcal{\ Q}_{a}$ and ${\ \mathcal{Q}}_{t}^{'}$, as illustrated in Eq.~\ref{eq:12},
\begin{equation}
\label{eq:12}
\revised{\mathcal{Z}_{a} = \text{Norm}\left(\left[{\ \mathcal{Q}}_{a} \mathcal{W}_{q},\ b_{a}\right]\right),
\mathcal{Z}_{t} = \text{Norm}({\ \mathcal{Q}}_{t}^{'}).}
\end{equation}
%<*response-r22-3>
\revised{Here, $\mathcal{Z}_{t} \in \mathbb{R}^{N \times C}$, $\mathcal{Z}_{a} \in \mathbb{R}^{M \times C}$ represent the multi-modal joint space embeddings, where $N$ is the number of trajectory queries matched with $\ \mathcal{Z}_{a}$ and $M$ is the number of collected audio queries.
$\mathcal{W}_{q}$ is a learnable matrix that transforms ${\ \mathcal{Q}}_{t}^{'}$ to feature dimension $C$, and $b_{a}$ is a learnable bias term.
Next, we compute the similarity matrix $\chi_{\text{sim}} $ through matrix multiplication as depicted in Eq.~\ref{eq:13},}
\begin{equation}
\label{eq:13}
    \revised{\chi_{\text{sim}} = \left[(\mathcal{Z}_{t} \mathcal{Z}_{a}^{\mathcal{T}}) / e^{\varphi}  ,\ b_{\rho}\right],}
\end{equation}
\revised{where $e^{\varphi}$ is a modulation factor and $\varphi$ is a learnable parameter. $b_{\rho}$ is a learnable regularization term.
Finally, we compute the focal loss for each element in $\chi_{\text{sim}}$. }
%</response-r22-3>
In summary, the loss of ACTL is expressed via Eq.~\ref{eq:14},
\begin{equation}
\label{eq:14}
\!\!\mathcal{L}_{act}(\chi_{\text{sim}})\!\!=\!\! \begin{cases}
\!- \frac{1}{\psi }\!\sum_{\chi_{\text{sim}}^{i,j} \in \psi}{(1 \!- \! \chi_{\text{sim}}^{i,j})^{\gamma }\log(\chi_{\text{sim}}^{i,j})}, \chi_{\text{sim}}^{i,j} \! \in \!\mathcal{P} \\
\!- \frac{1}{\psi}\!\sum_{\chi_{\text{sim}}^{i,j}\in \psi}{\!(\chi_{\text{sim}}^{i,j})^{\gamma }\!\log(1 \!- \!\chi_{\text{sim}}^{i,j})},\!\chi_{\text{sim}}^{i,j} \! \in \! \mathcal{N},
\end{cases}
\end{equation}
where $\mathcal{P}$ and $\mathcal{N}$ represent positive and negative referents in the ground truth, respectively. $\psi \in |\mathcal{P \cup N|}$ is the cardinality, and $\chi_{\text{sim}}^{i,j}$ represents the similarity between the $i$-th trajectory query and the $j$-th audio query.

\section{Established Benchmarks}
\label{sec:datasets}
\begin{figure*} [h]
\centering
\includegraphics[width=1.0\linewidth]{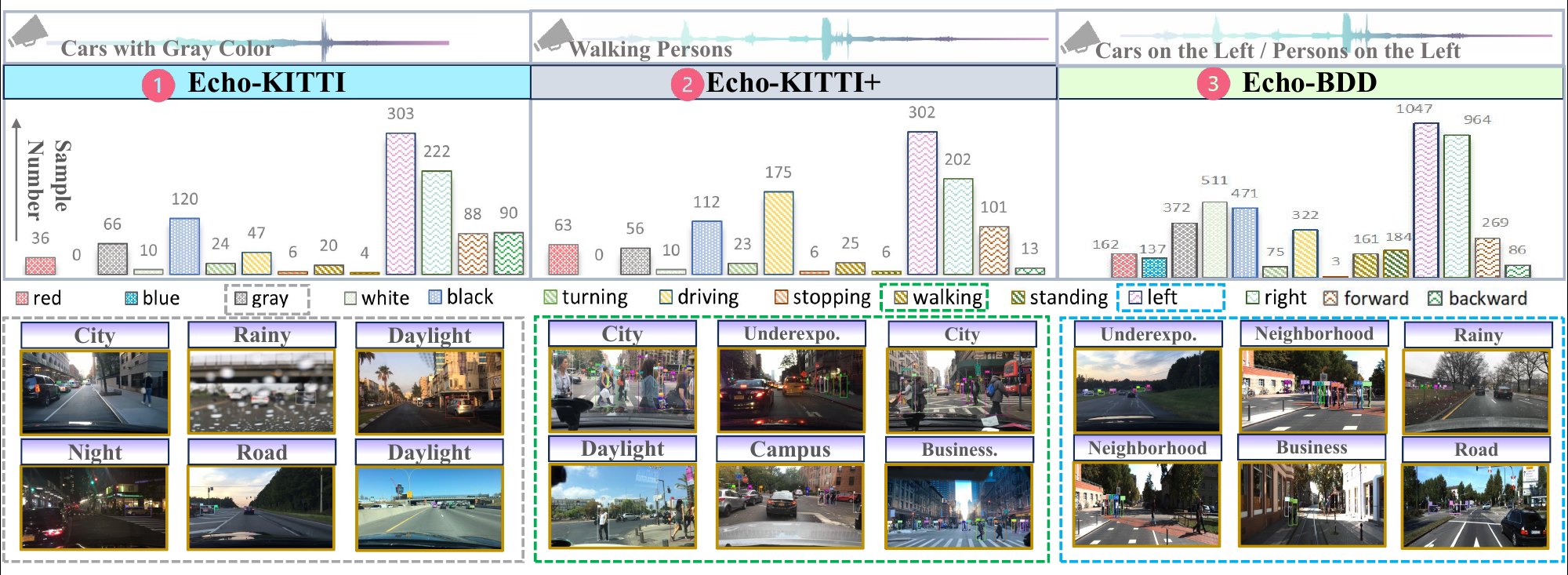}
\caption{\revised{Overview of the AR-MOT benchmarks. The video includes $4$ attributes, \textit{i.e.}, \textbf{Object}, \textbf{Scene}, \textbf{Weather}, and \textbf{Quality}, our benchmarks encompasses $6$ objects, \textit{i.e.}, \textit{Pedestrian}, \textit{Car}, \textit{Motorcycle}, \textit{Truck}, \textit{Bus}, \textit{Bicycle}, $5$ scenarios, \textit{i.e.}, \textit{Road}, \textit{City}, \textit{Neighborhood}, \textit{Business district}, and \textit{Campus}, $6$ weather conditions, \textit{i.e.}, \textit{Daylight}, \textit{Night}, \textit{Foggy}, \textit{Snowy}, \textit{Rainy}, and \textit{Cloudy}, $5$ video qualities, \textit{i.e.}, \textit{Normal}, \textit{Blur}, \textit{Overexposure}, \textit{Underexposure}, and \textit{Low-resolution}.
The audio expressions involve $4$ attributes of the objects, \textit{i.e.}, \textbf{Position}, \textbf{Color}, \textbf{Movement}, and \textbf{Gender}. It involves $4$ positions, \textit{i.e.}, \textit{Left}, \textit{Right}, \textit{Forward}, and \textit{Opposite}, $6$ colors, \textit{i.e.}, \textit{Red}, \textit{Black}, \textit{Blue}, \textit{Gray}, \textit{Yellow}, and \textit{White}, $5$ motions, \textit{i.e.}, \textit{Turning}, \textit{Driving}, \textit{Stopping}, \textit{Walking}, and \textit{Standing}, $2$ genders, \textit{i.e.}, \textit{Male} and \textit{Female}.}}
\label{fig:dataset}
\end{figure*}

\begin{table*}[h!]
\centering
%<*response-r16-tab>
\setlength\tabcolsep{6.0pt}
\renewcommand\arraystretch{1.0}
\caption{Comparison of AR-MOT benchmarks with existing video-based datasets. \XSolidBrush means unavailable, while \Checkmark denotes the opposite. Weather and Quality mean the types of weather and image quality, respectively. Note that the audio in AVSBench~\cite{zhou2022audio}, Flickr-SoundNet~\cite{senocak2018learning}, and VGG-SS~\cite{chen2021localizing} is the natural sound emitted by the object itself in the video. \revised{Bbox indicates whether the dataset has annotation bounding boxes.}}
\resizebox{\linewidth}{!}{\begin{tabular}{lccccccccccc}
\toprule [2pt]
\multirow{2}{*}{Dataset}             & \multirow{2}{*}{Year} & \multirow{2}{*}{Publication} & \multicolumn{5}{c}{Annotation}  & \multirow{2}{*}{Audio Number} & \multirow{2}{*}{Text Number} & \multirow{2}{*}{\makecell[c]{Instances \\ per-expression}} & \multirow{2}{*}{\makecell[c]{Instances \\ per-video}}\\
\cline{4-8}
 & &  & Bbox & Audio  & Text & Weather & Quality\\
\toprule [1pt]
A2D~\cite{gavrilyuk2018actor}                 & $2018$ & CVPR       & \XSolidBrush & \XSolidBrush     &  \Checkmark           & \XSolidBrush  & \XSolidBrush & \XSolidBrush & $\revised{6,656}$ & $1$                        & $1$       \\
J-HMDB~\cite{gavrilyuk2018actor}              & $2018$ & CVPR       & \XSolidBrush   & \XSolidBrush     & \Checkmark    & \XSolidBrush  & \XSolidBrush & \XSolidBrush & $928$ & $1$                        & $1$                 \\
RVOS~\cite{seo2020urvos}   & $2020$ & ECCV       & \XSolidBrush & \XSolidBrush & \Checkmark & \XSolidBrush & \XSolidBrush  & \XSolidBrush   & $15,009$  & $1$                        & $1$           \\
MeViS~\cite{ding2023mevis}               & $2023$ & ICCV      & \XSolidBrush & \XSolidBrush & \Checkmark & \XSolidBrush & \XSolidBrush  & \XSolidBrush  & $28,570$  & $1.59$                     & $4.28$   \\
Refer-KITTI~\cite{wu2023referring}         & $2023$ & CVPR       & \Checkmark   & \XSolidBrush & \Checkmark & \XSolidBrush & \XSolidBrush  & \XSolidBrush   & $818$ & $10.7$                     & $10.5$            \\

\toprule [1pt]
Flickr-SoundNet~\cite{senocak2018learning} & $2018$ & CVPR       & \Checkmark & \Checkmark  & \XSolidBrush & \XSolidBrush & \XSolidBrush  & $11,849$   & \XSolidBrush & $1$                        & $1$  \\
VGG-SS~\cite{chen2021localizing} & $2021$ & CVPR       & \Checkmark & \Checkmark  & \XSolidBrush & \XSolidBrush & \XSolidBrush  & $\revised{5,158}$   & \XSolidBrush  & $1$                        & $1$              \\
A-RVOS~\cite{pan2022wnet} & $2022$ & CVPR       & \XSolidBrush & \Checkmark  & \XSolidBrush & \XSolidBrush & \XSolidBrush  & $11,226$   & \XSolidBrush    & $1$                        & $1$               \\
AVSBench~\cite{zhou2022audio}            & $2022$ & ECCV       & \XSolidBrush & \Checkmark  & \XSolidBrush & \XSolidBrush & \XSolidBrush  & $10,852$   & \XSolidBrush  & $1$                        & $1$              \\
A-A2D~\cite{pan2022wnet}               & $2022$ & CVPR       & \XSolidBrush & \Checkmark  & \XSolidBrush & \XSolidBrush & \XSolidBrush  & $\revised{6,656}$   & \XSolidBrush  & $1$                        & $1$               \\
A-J-HMDB~\cite{pan2022wnet}            & $2022$ & CVPR       & \XSolidBrush & \Checkmark  & \XSolidBrush & \XSolidBrush & \XSolidBrush  & $929$   & \XSolidBrush    & $1$                        & $1$             \\
\toprule [1pt]
\rowcolor[gray]{.9} 
\textbf{AR-MOT} (ours)   & $2024$ & $-  $         & \Checkmark   & \Checkmark & \Checkmark & \Checkmark & \Checkmark  & $\revised{6,220}$  & $\revised{6,220}$    & $14.1$                     & $87.4$               \\
\toprule [2pt]
\end{tabular}}
%</response-r16-tab>
\label{sup:lab8}
\end{table*}

\subsection{Auditory Referring MOT Benchmarks} 
\subsubsection{Motivation of AR-MOT Benchmarks} Contemporary studies exploring audio in referring scenes understanding have greatly advanced the field, yet they have associated limitations in existing benchmarks, 
\textit{i.e.}, AVOS~\cite{pan2022wnet}, AVSBench~\cite{zhou2022audio}, Flickr-SoundNet~\cite{senocak2018learning}, and VGG-SS~\cite{chen2021localizing}. As shown in Table~\ref{sup:lab8}, these audio-based referring scenes understanding datasets often pair one audio expression with one visual object, diverging from real-world scenarios where one audio expression may correspond to multiple visual objects or vice versa. Moreover, these datasets exhibit simplicity in both audio expressions and scenes, lacking representation of the complexity and variability (\textit{i.e.}, weather and quality) inherent in real-world situations. 
Consequently, the inadequacy of these audio-based referring scene understanding datasets hinders the accurate assessment of scenarios with multiple referring objects, intricate expressions, and diverse conditions.

To foster the development of AR-MOT, we construct the AR-MOT benchmarks enriching KITTI~\cite{geiger2012we,wu2023language} and BDD100K~\cite{yu2020bdd100k} with text and extensive audio annotations.
These two datasets are selected due to their diverse images encompassing scenes, object classes, weather conditions, and image quality. In particular, BDD100K stands out for its richness, offering a foundation for simulating human-vehicle interactions in real driving scenarios.

\subsubsection{Annotation Rules} We first delineate $3$ fundamental attributes and gauge the complexity of linguistic expressions from the perspective of in-vehicle sensors as shown in Fig.~\ref{fig:dataset}. The foundational attributes comprise the \textbf{Position}, \textbf{Color}, and \textbf{Movement} of the object, while the complexity of expressions with different speaking speeds and noise tolerances is categorized into \textbf{Short}, \textbf{Medium}, and \textbf{Long}. Second, we formulate $4$ fundamental attributes of the image grounded in real-road conditions, \textit{i.e.}, \textbf{Object}, \textbf{Scene}, \textbf{Weather}, and \textbf{Quality}. Subsequent to these attribute specifications, we filter $68$ eligible videos from the KITTI and BDD100K datasets. Specifically, $18$ videos are sourced from KITTI, while the remaining are derived from BDD100K.
%<*response-r22-4>
\revised{Then, we build a \textit{language attribute library} based on the attributes of \textbf{Class}, \textbf{Position}, \textbf{Color}, \textbf{Movement}, and \textbf{Gender}. We then obtain the \textit{frame IDs} of each object by using the ground truth of KITTI and BDD100K. Next, we construct the fundamental \textit{object-language relationship library}, denoted as $\mathcal{D} =\{\textit{language attribute library}, \textit{frame IDs}\}$.
Based on $\mathcal{D} $, we curate the corpus by using the key attributes in \textit{frame IDs} as the sentence backbone and incorporating object information from the images.}
%</response-r22-4>
To ensure corpus diversity, we stipulate the inclusion of each object in no fewer than $4$ synonymous sentences, with varying levels of complexity for every linguistic expression and different speaking speeds and noise tolerances. Moreover, we employ $4$ speakers ($2$ males and $2$ females) to orally deliver sentences from the corpus. To ensure high-quality audio annotations, readers are instructed to articulate proficiently, avoiding stuttering or interruptions. Furthermore, we conduct double-checks on the AR-MOT benchmarks to guarantee the accuracy of both the text and audio files.

\subsubsection{Dataset Statistics and Split} Based on the design principles, the three benchmarks are established, and the details of them are described as follows:

\textbf{1) Echo-KITTI:} Echo-KITTI contains a total of $18$ videos and $820$ audio expressions, with an average expression length and an average number of objects per video of $4.5$ and $10.5$, respectively.
In our experimental setup, we allocate $15$ out of the $18$ individual videos for the training set, reserving the remaining $3$ for evaluation within the test set.

\textbf{2) Echo-KITTI+:} Derived from Echo-KITTI, Echo-KITTI+ is an expansion where the average expression length has been augmented from \marked{$4.5$ to $8.8$}. The audio expression of Echo-KITTI is ``counter-direction cars on the left'', but it can be extended to ``cars traveling in the opposite direction are situated on the left side'' on the Echo-KITTI+ dataset. Echo-KITTI+ is designed to evaluate the influence of distinct levels of expression difficulty on the AR-MOT. All settings remain congruent with the Echo-KITTI.

\textbf{3) Echo-BDD:} Echo-BDD dataset has a uniform distribution of expressions, the most comprehensive coverage of scenes, and the largest number of objects covered in the current referring MOT task. It contains a total of $50$ videos, with more than an \marked{average of $120$ objects} per video.
We take $42$ videos out of $50$ for training and the remaining $8$ as the test set.

\section{Experiments}
\label{sec:experiment}

\begin{table*}[]
\centering
\renewcommand\arraystretch{1.2}
\setlength\tabcolsep{4.0pt}
\caption{Quantitative results on the Echo-KITTI, Echo-KITTI+, and Echo-BDD datasets.
Note that the TransRMOT~\cite{wu2023referring} uses HuBERT to replace BERT. \textbf{Bold} and \underline{underlined} results indicate the best performing and second-best performing results, respectively.}
\resizebox{\linewidth}{!}{\begin{tabular}{l|ccccc|ccccc|ccccc}
\toprule [2pt]
& \multicolumn{5}{c|}{Echo-KITTI}   & \multicolumn{5}{c|}{Echo-KITTI+}  & \multicolumn{5}{c}{Echo-BDD}     \\
\cline {2-16}
\multicolumn{1}{l|}{\multirow{-2}{*}{Method}}      & HOTA$\uparrow$  & DetA$\uparrow$  & AssA$\uparrow$  & MOTA$\uparrow$  & IDF1$\uparrow$  & HOTA$\uparrow$  & DetA$\uparrow$  & AssA$\uparrow$  & MOTA$\uparrow$  & IDF1$\uparrow$  & HOTA$\uparrow$  & DetA$\uparrow$  & AssA$\uparrow$  & MOTA$\uparrow$  & IDF1$\uparrow$  \\
\midrule [1pt]
TransTrack~\cite{sun2020transtrack}  &   $29.16$   &  $17.62$ & $48.86$  &  $6.60$ &  $31.24$ & $27.30$ &	$15.32$	& $49.48$      &  $7.58$  &  $29.50$  &  $16.90$		
   &   $8.35$   &  $34.75$    & $1.67$ &  $13.47$ \\
TrackFormer~\cite{meinhardt2022trackformer}  &  $30.78$    &   $18.73$   &  $51.77$    &   $7.38$   &   $31.96$   & $29.93$  & $18.98$  & $48.61$  &     $8.36$ & $34.19$  & $17.32$  &   $11.56$   &  $26.37$    &   $2.92$   &   $15.89$   \\
CO-MOT~\cite{yan2023bridging}    &   $30.45$   &  $17.57$ & $\underline{53.92}$  & $6.64$ & $31.51$ &   $28.32$   &  $16.00$    &   $\underline{51.23}$   &   $7.02$   &   $28.32$      &  $20.88$    &  $13.78$    &  $31.95$    &   $4.76$   &  $19.98$  \\
MeMOTR~\cite{gao2023memotr}      &  $\underline{33.88}$ & $20.65$ & $\mathbf{56.83}$ &  $8.62$ & $38.44$ &  $\underline{33.57}$    &  $21.24$   &  $\textbf{54.31}$ &   $7.02$    &   $37.67$   &   $22.53$   &  $11.10$ &  $\underline{46.30}$ &  $5.21$  & $19.15$  \\
MOTRv2~\cite{zhang2023motrv2}      &  $33.69$    &   $\underline{24.00}$   &   $48.82$   &   $\underline{10.97}$   &  $38.70$    & $31.64$  &  $20.12$ & $51.14$ & $8.35$ & $35.67$ &  $24.60$ &  $11.94$   &     $\textbf{51.60}$ &  $\underline{5.41}$  &  $20.22$     \\
TransRMOT~\cite{wu2023referring} &      $33.58$ & $23.81$ &  $48.78$  &  $8.69$  &  $\underline{39.06}$   & $32.78$  & $\underline{24.71}$  & $44.83$  & $\underline{10.35}$  &  $\underline{37.88}$  & $\underline{28.29}$ &  $\underline{19.71}$   &     $42.65$ &  $5.38$  &  $\underline{30.70}$  \\
\rowcolor[gray]{.9} 
\textbf{EchoTrack} (ours)   & $\mathbf{37.14}$     &  $\mathbf{27.39}$    &  $51.88$    &  $\mathbf{13.41}$    &  $\mathbf{44.30}$    &  $\mathbf{36.59}$    &   $\mathbf{27.01}$  &  $51.01$   &  $\mathbf{16.63}$  & $\mathbf{44.74}$ &  $\mathbf{31.12}$ &  $\mathbf{21.96}$ & $44.62$ & $\mathbf{9.57}$ & $\mathbf{31.78}$ \\
\bottomrule [2pt]
\end{tabular}}
\label{lab:lab7}
\end{table*}

%<*response-r110-2>
\revised{In this section, we first introduce the datasets and evaluation metrics we used in Sec.~\ref{datasets}. Then, we present the implementation details in Sec.~\ref{details}. In Sec.~\ref{benchmarks}, a comparison experiment is made between EchoTrack and other existing methods when training on AR-MOT benchmarks. Subsequently, a generalizability analysis is present in Sec.~\ref{gener}. Finally, parameter analysis and ablation experiments are performed to check the effectiveness of Bi-FCFM and ACTL in Sec.~\ref{ablation}.}
%</response-r110-2>

\subsection{Datasets and Evaluation Metrics} \label{datasets}

\textbf{Datasets.} 
%<*response-r27>
\revised{We use six datasets, Echo-KITTI, Echo-KITTI+, Echo-BDD, Refer-KITTI~\cite{wu2023referring}, Refer-KITTI+, and Refer-BDD, to verify the tracking performance of our proposed EchoTrack. Refer-KITTI is a well-recognized dataset in RMOT tasks, Refer-KITTI+ and Refer-BDD are the text-based Echo-KITTI+ and Echo-BDD datasets, respectively.}
%</response-r27>

\textbf{Evaluation Metrics.} To evaluate the AR-MOT task, we formulate the evaluation metrics, drawing inspiration from MOT~\cite{yan2023bridging,gao2023memotr,zhang2023motrv2}, and RMOT~\cite{wu2023referring,wu2023language, nguyen2023type}.
We first employ HOTA~\cite{luiten2021hota} as a benchmark for calculating the similarity between the predicted trajectory and the ground-truth trajectory. Additionally, we utilize the metrics DetA and AssA to evaluate the performance of AR-MOT. 
%DetA and AssA correspond to the scores associated with the IoU for detection and association, respectively. 
Furthermore, we incorporate $\mathrm{MOTA}$ and $\mathrm{IDF1}$ following the~\cite{nguyen2023type, milan2016mot16, bernardin2008evaluating} to evaluate the effectiveness of AR-MOT. In addition, the metrics HOTA, DetA, AssA, MOTA, and IDF1 are calculated by the average value of different audio queries.
%<*response-r19-2>
\revised{Except for the metrics we used in AR-MOT, we also incorporate DetRe, DetPr, AssRe, AssPr, and LocA metrics to evaluate the tracking performance of the RMOT task following existing works~\cite{wu2023referring,du2023ikun,he2024visual}.}
%</response-r19-2>

\subsection{Implementation Details} \label{details}

Under deformable DETR style~\cite{DeformableDETR}, our transformer encoders and decoders are configured with $6$ layers.
The query count is set to $300$.
All experiments are performed on four NVIDIA GeForce RTX A6000 GPUs, with a batch size of $1$ per GPU.
For the training process, we randomly sample $2$ frames and $1$ audio queries on each occasion.
The number of training epochs is $100$. We use AdamW~\cite{loshchilov2018decoupled} with the initial learning rate of $1{\times}10^{-4}$, which is dropped by a factor of $10$ at the 50-th epoch. 
%<*response-r12-71>
\revised{The loss coefficients are set as $2$, $2$, $5$, and $2$ for $\mathcal{L}_{cls}$, $\mathcal{L}_{iou}$, $\mathcal{L}_{l_{1}}$, and $\mathcal{L}_{act}$, respectively. In the evaluation phase, a referred object is obtained when the class score surpasses $0.7$ and the referring score exceeds $0.5$.}
%</response-r12-71>

For other baseline models, \textit{e.g.}, TransTrack~\cite{sun2020transtrack}, TrackFormer~\cite{meinhardt2022trackformer}, MOTRv2~\cite{zhang2023motrv2}, MeMOTR~\cite{gao2023memotr}, and CO-MOT~\cite{yan2023bridging}, we employ the incorporation of Bi-XAtt~\cite{yan2023universal} and refer loss\cite{wu2023referring} akin to EchoTrack, to tailor them for the AR-MOT task.
To generate detection proposals for MOTRv2, the YOLOX~\cite{ge2021yolox} detector is correspondingly trained for $20$ epochs, with a batch size set to $48$. 
When incorporating Bi-FCFM and ACTL into RMOT methods, \textit{e.g.} TransRMOT~\cite{wu2023referring}. We maintain consistency in parameter settings with those of the RMOT comparison methods. 

\subsection{Analysis on AR-MOT Benchmarks} \label{benchmarks}

\subsubsection{Comparison with State-of-the-Art AR-MOT Methods} We begin our analysis by assessing the overall performance of existing MOT methods on the AR-MOT benchmarks, as summarized in Table~\ref{lab:lab7}. 
To evaluate the generalizability of diverse tracking frameworks, we validate their performance across three distinct benchmarks: Echo-KITTI, Echo-KITTI+, and Echo-BDD. 
Specifically, we employ detection-based MOT frameworks, including MOTRv2~\cite{zhang2023motrv2}, end-to-end transformer-based methods, \textit{i.e.}, CO-MOT~\cite{yan2023bridging} and TrackFormer~\cite{meinhardt2022trackformer}, as well as MOTR-based MOT frameworks, \textit{i.e.}, MeMOTR~\cite{gao2023memotr} and TransRMOT~\cite{wu2023referring}.
The experimental results across various MOT frameworks reveal that the end-to-end MOTR-based framework surpasses other transformer-based methods in AR-MOT. 
This superiority is evident through an averaged improvement of $3.59$, $4.15$, and $6.77$ in HOTA across the three benchmarks, respectively.

\begin{table}[t!]
\setlength\tabcolsep{8pt}
\renewcommand\arraystretch{1.0}
\caption{Evaluation of our Bi-FCFM compared against state-of-the-art attention-based fusion modules trained on Echo-KITTI.}
\resizebox{\linewidth}{!}{\begin{tabular}{l|ccccc}
\toprule [2pt]
Method      & HOTA$\uparrow$  & DetA$\uparrow$  & AssA$\uparrow$  & MOTA$\uparrow$  & IDF1$\uparrow$  \\
\midrule [1pt]
Vanilla-CAtt~\cite{wu2023referring}    &  $\underline{34.06}$   & $23.17$ & $\underline{51.44}$  &  $\underline{12.42}$  &   $39.47$     \\
Bi-XAtt~\cite{yan2023universal} & $33.44$ & $25.24$ & $44.31$ & $11.10$ &   $37.67$   \\
DWT-CAtt~\cite{pan2022wnet}    &      $21.33$ &  $10.13$& $45.79$  & $-2.33$  &  $19.38$    \\
FFT-CAtt~\cite{miao2023spectrum}    &   $33.66$   & $\underline{26.92}$ &  $43.08$ & $8.61$  &  $\underline{39.59}$  \\
\midrule [1pt]
\rowcolor[gray]{.9} 
\textbf{Bi-FCFM} (ours)      & $\mathbf{37.14}$     &  $\mathbf{27.39}$    &  $\mathbf{51.88}$    &  $\mathbf{13.41}$    &  $\mathbf{44.30}$        \\
\bottomrule [2pt]
\end{tabular}}
\label{lab:label_bifcfm}
\end{table}

\begin{figure*}
\centering
\includegraphics[width=0.85\textwidth]{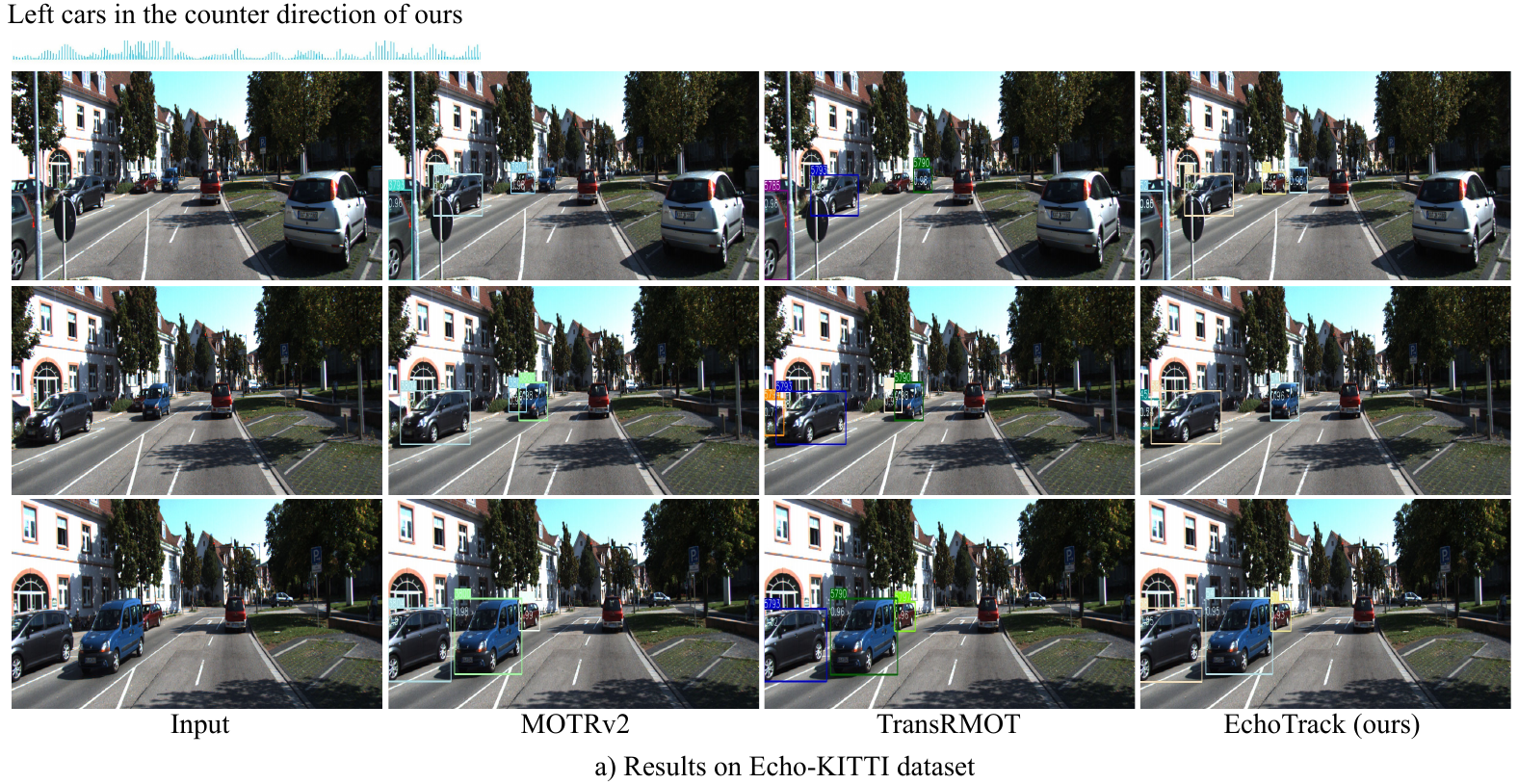}

\includegraphics[width=0.85\textwidth]{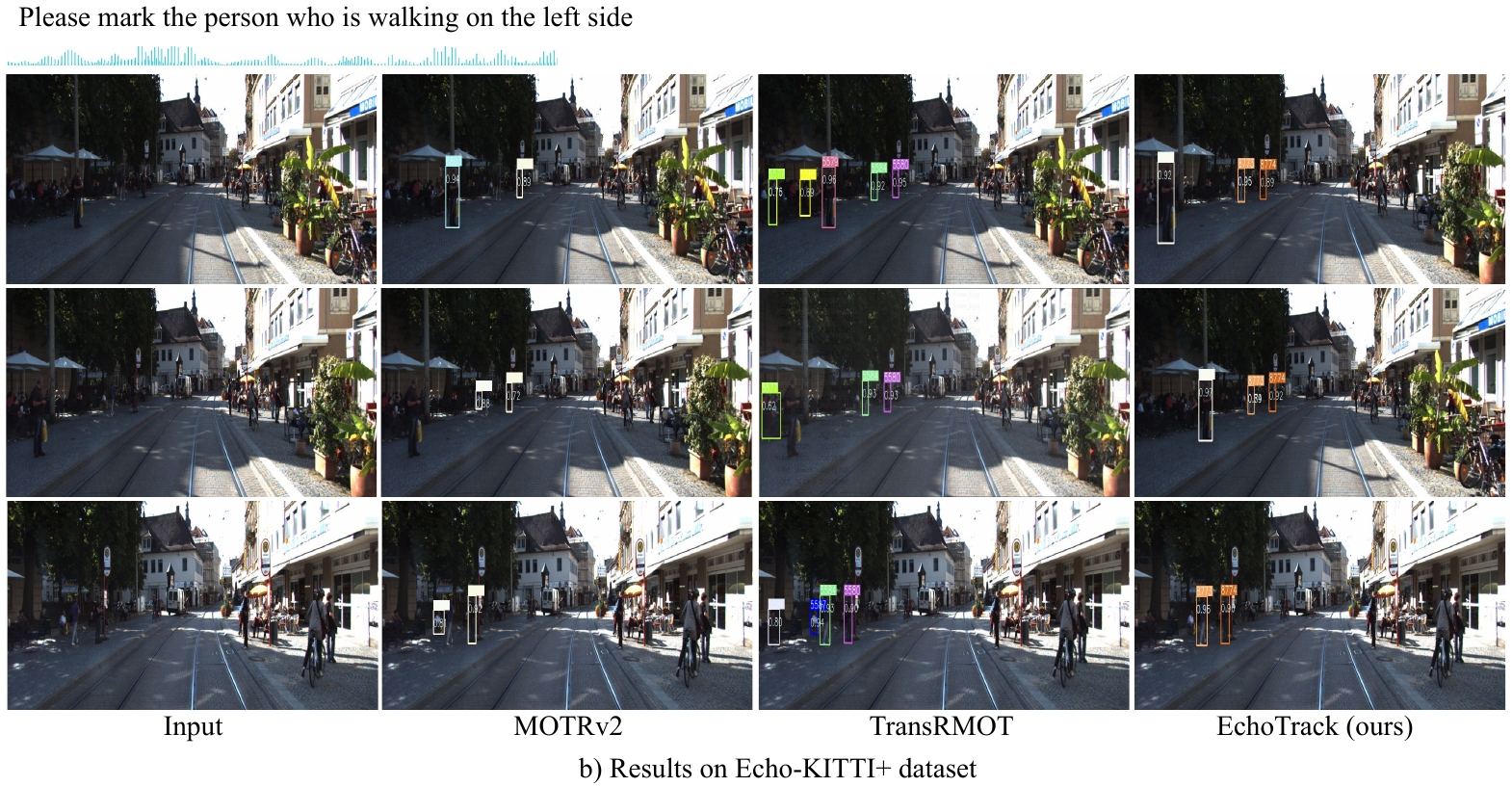}

\includegraphics[width=0.85\textwidth]{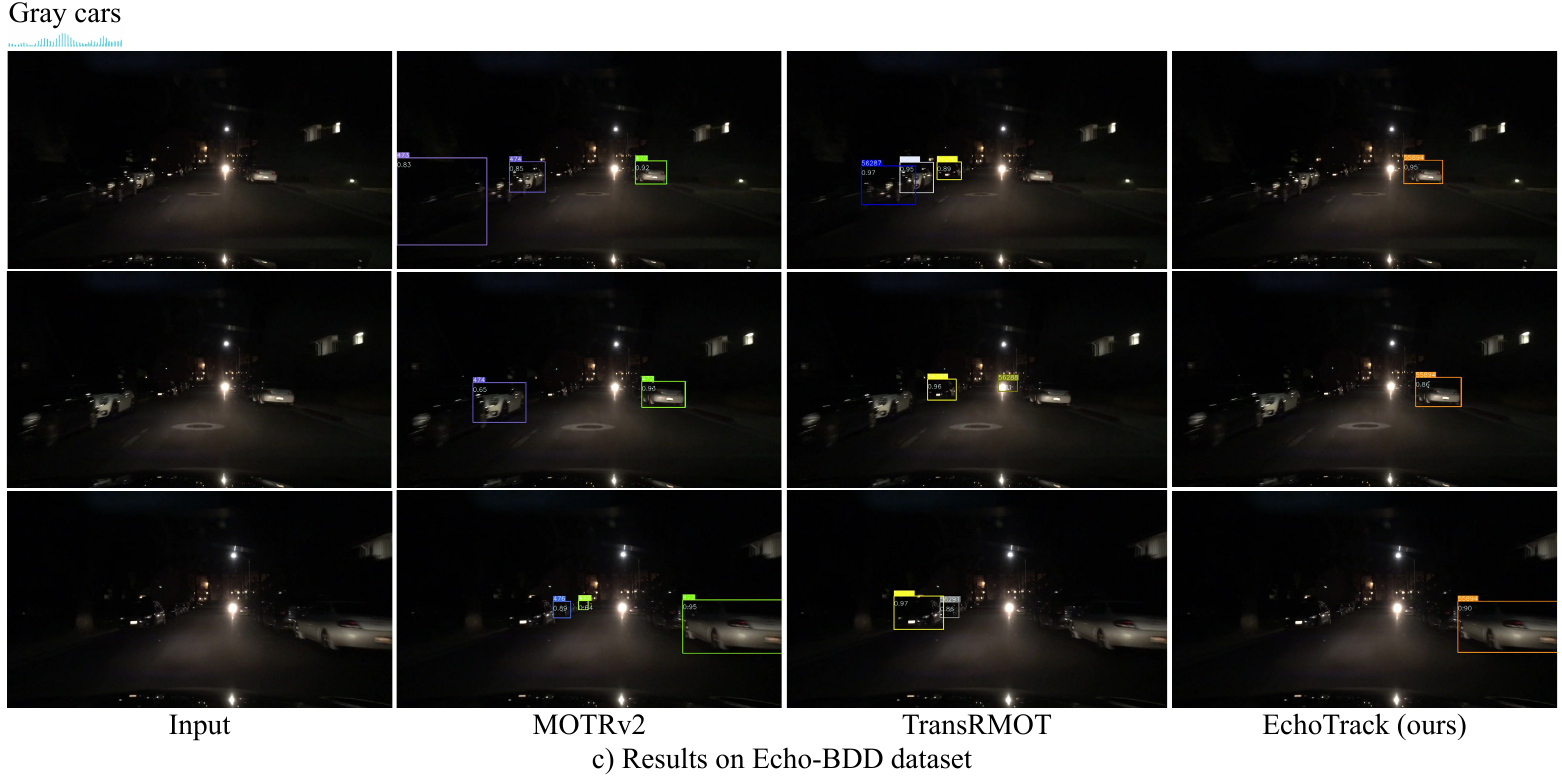}
\caption{Qualitative results of different state-of-the-art methods include MOTRv2~\cite{zhang2023motrv2}, TransRMOT~\cite{wu2023referring}, and the proposed EchoTrack on Echo-KITTI, Echo-KITTI+, and Echo-BDD datasets.  EchoTrack shows leading tracking performance.}
\label{sup:fig1}
\end{figure*}

\begin{figure} [t!]
\centering
\includegraphics[width=1.0\linewidth]{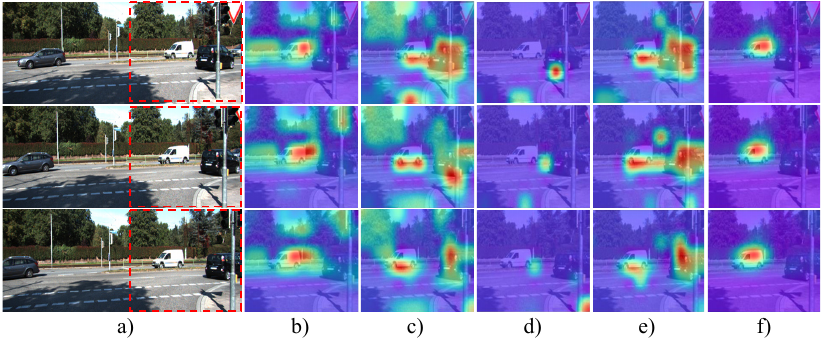}\caption{The tracking heatmaps of different fusion modules of the audio query is ``cars-in-light-color'', from left to right, are: a) Input b) Vanilla-CAtt~\cite{wu2023referring}, c) Bi-XAtt~\cite{yan2023universal}, d) DWT-CAtt~\cite{pan2022wnet}, e) FFT-CAtt~\cite{miao2023spectrum}, and f) proposed Bi-FCFM. Bi-FCFM shows leading focus performance.}
\label{fig:heatmap}
\end{figure}

\begin{table*}[]
\centering
%<*response-r19-tab1>

\renewcommand\arraystretch{1.0}
\setlength\tabcolsep{5pt}
\caption{Quantitative results on Refer-KITTI~\cite{wu2023referring}. Note that EchoTrack$_{\text{RMOT}}$ means EchoTrack with BERT is used for the RMOT task. \revised{Following~\cite{wu2023referring,du2023ikun}, the HOTA, DetA, AssA, DetRe, DetPr, AssRe, AssPr, LocA, MOTA, and IDF1 are reported.  $^\ast$ means that use the frame correction strategy. The improvements calculated based on corresponding baseline methods are in \improved{Green}.}}
\resizebox{\linewidth}{!}{\begin{tabular}{l|llllllllll}
\toprule [2pt]
\multicolumn{1}{l|}{Method}   & HOTA$\uparrow$    & DetA$\uparrow$   & AssA$\uparrow$ & DetRe$\uparrow$ & DetPr$\uparrow$ & AssRe$\uparrow$ & AssPr$\uparrow$ & LocA$\uparrow$  & \revised{MOTA$\uparrow$}  & \revised{IDF1$\uparrow$} \\
\midrule [1pt]

DeepSORT~\cite{wojke2017simple}     & $25.59$&$19.76$&$34.31$&$26.38$&$36.93$&$39.55$&$61.05$&$71.34$&- &-  \\
ByteTrack~\cite{zhang2022bytetrack} & $\,24.95$&$15.50$&$43.11$&$18.25$&$43.48$&$48.64$&$70.72$&$73.90$&-&- \\
CStrack~\cite{liang2022rethinking}   &$\,27.91$&$20.65$&$39.00$&$33.76$&$32.61$&$43.12$&$71.82$&$79.51$&-&-\\
TransTrack~\cite{sun2020transtrack}   &$\,32.77$&$23.31$&$45.71$&$32.33$&$42.23$&$49.99$&$78.74$&$79.48$&-&- \\
TrackFormer~\cite{meinhardt2022trackformer} &$\,33.26$&$25.44$&$45.87$ &$35.21$&$42.19$&$50.26$&$78.92$&$79.63$&-&-\\
\revised{DeepRMOT~\cite{he2024visual}}  & $\,\revised{\mathbf{39.55}}$ & $\revised{\underline{30.12}}$ & $\revised{\mathbf{53.23}}$ & $\revised{\underline{41.91}}$ & $\revised{\underline{47.47}}$ & $\revised{\mathbf{58.47}}$ & $\revised{\mathbf{82.16}}$ & $\revised{\mathbf{80.49}}$ & \revised{-} &  \revised{-}  \\
TransRMOT~\cite{wu2023referring}   &$\,38.06$ &$29.28$  &$50.83$    &$40.19$ 
 &$47.36$ & $55.43$  &$\underline{81.36}$  &$\underline{79.93}$ &$\underline{\revised{9.03}}$  &$\underline{\revised{46.40}}$ \\
\rowcolor[gray]{.9} 
\textbf{EchoTrack}$^{\text{RMOT}}$ (ours)  & $\underline{39.47}_{\improved{+ 1.41}}$ & $\mathbf{31.19}_{\improved{+ 2.11}}$ & $\underline{51.56}_{\improved{+ 0.73}}$ & $\mathbf{42.65}_{\improved{+ 2.46}}$ & $\mathbf{48.86}_{\improved{+ 1.50}}$ & $\underline{56.68}_{\improved{+ 1.25}}$ & $81.21_{\improved{- 0.15}}$ & $\underline{79.93}_{\improved{+ 0.00}}$ & $\revised{\mathbf{11.20}}_{\improved{+ 2.17}}$ & $\revised{\mathbf{47.43}}_{\improved{+ 1.03}}$  \\
\midrule [1pt]
\revised{iKUN$^{\ast}$}~\cite{du2023ikun}   &\revised{$\,\underline{48.84}$} &\revised{$35.74$}  &\revised{$\mathbf{66.80}$} & $\revised{\underline{51.97}}$ & $\revised{52.26}$ & $\revised{\mathbf{72.95}}$ & $\revised{87.09}$ & - &\revised{$12.26$}  &\revised{$\underline{54.05}$} \\
\revised{TransRMOT$^{\ast}$}~\cite{wu2023referring} &\revised{$\,46.56$} &\revised{$\underline{37.97}$}  &\revised{$57.33$} & \revised{$49.69$} & \revised{$\underline{60.10}$} & \revised{$61.02$} & \revised{$\mathbf{89.67}$} & \revised{$\underline{90.33}$} &\revised{$\underline{24.68}$}  &\revised{$53.85$} \\
\rowcolor[gray]{.9} 
\revised{\textbf{EchoTrack}$^{\text{RMOT}^{\ast}}$ (ours)} & $\mathbf{\revised{48.86}}_{\improved{+ 2.30}}$ & $\revised{\mathbf{41.26}_{\improved{+ 3.29}}}$ & $\revised{\underline{57.59}_{\improved{+ 0.26}}}$ & $\revised{\mathbf{53.42}}_{\improved{+ 3.73}}$ & $\revised{\mathbf{62.83}}_{\improved{+ 2.73}}$ & $\revised{\underline{61.61}}_{\improved{+ 0.59}}$ & $\revised{\underline{89.33}}_{\improved{- 0.34 }}$ & $\revised{\mathbf{90.74}}_{\improved{+ 0.42}}$ & $\revised{\mathbf{29.45}_{\improved{+ 4.77}}}$ & $\revised{\mathbf{55.94}_{\improved{+ 2.09}}}$  \\
\bottomrule [2pt]
\end{tabular}}

%</response-r19-tab1>
\label{lab:label2}
\end{table*}

\begin{table*}[]
\centering
%<*response-r19-tab2>
\renewcommand\arraystretch{1.0}
\setlength\tabcolsep{5pt}
\caption{Quantitative results on Refer-KITTI+ and Refer-BDD datasets. \revised{Following~\cite{wu2023referring,du2023ikun}, we incorporate the HOTA, DetA, AssA, DetRe, DetPr, AssRe, AssPr, LocA, MOTA, and IDF1 to evaluate the proposed Bi-FCFM and ACTL used in the RMOT task.}}
\resizebox{\linewidth}{!}{\begin{tabular}{l|llllllllll}
\toprule [2pt]
Method      & HOTA$\uparrow$    & DetA$\uparrow$   & AssA$\uparrow$ & DetRe$\uparrow$ & DetPr$\uparrow$ & AssRe$\uparrow$ & AssPr$\uparrow$ & LocA$\uparrow$  & \revised{MOTA$\uparrow$}  & \revised{IDF1$\uparrow$}  \\
\midrule [1pt]

\multicolumn{8}{l}{\textbf{\revised{\textit{Refer-KITTI+ dataset}}}} \\
TransRMOT~\cite{wu2023referring}   & $\underline{35.32}$    &$\underline{25.61}$ 
& $\underline{50.33}$  &$\mathbf{40.05}$  &$\underline{38.45}$ 
 &$\mathbf{55.40}$   &$\underline{81.23}$  &$\underline{79.44}$ & $\underline{\revised{5.59}}$   &$\underline{\revised{40.99}}$   \\
\rowcolor[gray]{.9} 
\textbf{EchoTrack}$^{\text{RMOT}}$ (ours)  & $\mathbf{37.46}_{\improved{+ 2.14}}$ & $\mathbf{28.83}_{\improved{+ 3.22}}$ & $\mathbf{50.39}_{\improved{+ 0.06}}$ & $\underline{39.83}_{\improved{- 0.22}}$ & $\mathbf{46.70}_{\improved{+ 8.25}}$ & $\underline{54.14}_{\improved{- 1.26}}$ & $\mathbf{82.57}_{\improved{+ 1.34}}$ & $\mathbf{79.97}_{\improved{+ 0.53}}$ & $\mathbf{\revised{7.36}}_{\improved{+ 1.77}}$ & $\mathbf{\revised{44.24}}_{\improved{+ 3.25}}$  \\
\midrule [1pt]
\multicolumn{8}{l}{\textbf{\revised{\textit{Refer-BDD dataset}}}} \\ 
\revised{MOTRv2~\cite{zhang2023motrv2}}   & \revised{$35.35$} & \revised{$22.88$}  & \revised{$\mathbf{55.24}$}  & \revised{$\mathbf{46.35}$} & \revised{$30.33$} & \revised{$\mathbf{64.19}$} & \revised{$75.47$}  & \revised{$85.3$}  & \revised{$\underline{4.77}$} & \revised{$38.87$}\\
\revised{CO-MOT~\cite{yan2023bridging}}   & \revised{$\underline{36.72}$} & \revised{$24.88$}  & \revised{$\underline{54.73}$}  & \revised{$30.01$} & \revised{$\mathbf{56.95}$} & \revised{$\underline{60.04}$} & \revised{$\underline{82.87}$}  & \revised{$\underline{86.50}$}  & \revised{$3.51$} & \revised{$39.76$}\\
TransRMOT~\cite{wu2023referring}    & $34.79$ &$\underline{26.22}$  &$47.56$ & $\underline{38.46}$    &$41.54$  &$51.35$   &$81.76$  &$79.37$ & $\revised{3.04}$    &$\underline{\revised{40.60}}$ \\
\rowcolor[gray]{.9} 
\textbf{EchoTrack}$^{\text{RMOT}}$ (ours)  & $\mathbf{38.00}_{\improved{+ 3.21}}$ & $\mathbf{28.57}_{\improved{+ 2.35}}$  & $51.24_{\improved{+ 3.68}}$ & $36.60_{\improved{- 1.86}}$ & $\underline{54.25}_{\improved{+ 12.71}}$ & $55.32_{\improved{+ 3.97}}$ & $\mathbf{85.81}_{\improved{+ 4.05}}$ & $\mathbf{86.99}_{\improved{+ 7.62}}$ & $\mathbf{\revised{12.90}}_{\improved{+ 9.68}}$ & $\mathbf{\revised{41.74}}_{\improved{+ 1.14}}$  \\

\bottomrule [2pt]
\end{tabular}}

%</response-r19-tab2>

\label{sup:label1}
\end{table*}

Next, we direct our attention to examining the impact of the complexity of linguistic expressions on various MOT frameworks.
A comparative analysis of experimental results between Echo-KITTI and Echo-KITTI+ reveals fluctuations in the tracking ability of all models as the complexity of the language increases.
Notably, the end-to-end MOTR-based MOT framework exhibits more stable performance compared to the detection-based frameworks.

Subsequently, we delve into an analysis of the impacts of diverse scenes, complexities, and image quality on the MOT framework.
Notably, the more complex Echo-BDD dataset presents a formidable challenge for AR-MOT.
The averaged HOTA of all methods on the intricate Echo-BDD exhibits a decrease of $10.17$ and $8.83$, respectively, when compared to their performance on the other two datasets.
These experimental findings underscore the heightened challenge posed by the specially designed Echo-BDD, offering a rich data foundation for the ongoing advancements in AR-MOT.

In conclusion, the performance of existing methods in the AR-MOT remains sub-optimal. This is attributed to the limited exploration of frequency domain cues in audio-visual features and an overlook of the loss of long-range propagation of referring features. The tracking performance and stability of current methods within the established benchmarks are deemed insufficient. Hence, we integrate our Bi-FCFM and ACTL into the baseline. The results demonstrate that EchoTrack yields substantial improvements across various metrics. Specifically, it achieves averaged performance gains of $10.08$\% and $62.20$\% in HOTA and MOTA, respectively, compared to TransRMOT. 
Fig.~\ref{sup:fig1} provides more visualization results of MOTRv2, TransRMOT, and the proposed EchoTrack on the Echo-KITTI, Echo-KITTI+, and Echo-BDD datasets. While existing MOT methods cannot adequately address the diverse range of audio and video attributes present in those datasets, the EchoTrack equipped with Bi-FCFM and ACTL demonstrates superior performance on the challenging benchmarks.

\subsubsection{Comparison with State-of-the-Art Fusion Modules} Table~\ref{lab:label_bifcfm} presents the outcomes of Bi-FCFM in contrast to state-of-the-art fusion modules. 
Compared to DWT-CAtt~\cite{pan2022wnet} and Bi-XAtt~\cite{yan2023universal}, the proposed Bi-FCFM demonstrates superior performance. 
Even compared with the more advanced Vanilla-CAtt~\cite{wu2023referring} and FFT-CAtt~\cite{miao2023spectrum}, Bi-FCFM achieves at least $3.08$, $0.47$, $0.44$, $0.99$, and $4.71$ improvements in each of HOTA, DetA, AssA, MOTA, and IDF1. 
This indicates that our frequency domain adaptive interaction mechanism effectively enhances the effective learning of the tracker, providing more precise localization of the specified object. 
Fig.~\ref{fig:heatmap} visualizes the heatmaps of the different fusion modules. 
While existing fusion methods produce heatmaps that do not fully concentrate on the audio-specified object, our Bi-FCFM excels in precisely focusing on the specified object guided by audio cues.

\subsection{Generalizability Analysis} \label{gener}
\subsubsection{Application of EchoTrack to RMOT} 
We implement EchoTrack into text-based RMOT tasks to verify its efficacy.
The results on Refer-KITTI~\cite{wu2023referring} widely used in RMOT show that the proposed Bi-FCFM, as well as ACTL, can not only be applied to AR-MOT but also effectively improve the performance of the text-only RMOT task.
As shown in Table~\ref{lab:label2}, relative to TransRMOT, EchoTrack$_{\text{RMOT}}$ achieves consistent gains in all metrics.
Notably, the pivotal HOTA exhibits a notable increase of $3.93$, signifying an enhancement of $11.06$\%. 
%<*response-r19-1>
\revised{Additionally, we also provide results for the frame correction strategies. The experimental results show that the proposed EchoTrack can be seamlessly embedded into existing RMOT methods and bring consistent performance improvements to such methods. The proposed method exhibits competitive tracking performance with or without the use of a frame correction strategy.}
%</response-r19-1>
%<*response-r19>
Table~\ref{sup:label1} shows the results of \revised{Refer-KITTI+ and Refer-BDD datasets between EchoTrack and TransRMOT. We reach the following observations:
1) TransRMOT~\cite{wu2023referring} has limited tracking capabilities when applied to text-based AR-MOT benchmarks, whereas the proposed EchoTrack shows significant performance improvements. The HOTA results of TransRMOT across two datasets notably fall short in comparison to our EchoTrack, with particularly significant differences exhibited on Refer-BDD. Specifically, the results in HOTA achieved by the TransRMOT on Refer-BDD represent only $91.55\%$ of the performance delivered by the proposed EchoTrack.
2) The more complex Refer-BDD dataset poses a tougher challenge for the RMOT task. 
It can be seen that the average HOTA of TransRMOT~\cite{wu2023referring} on the more complex Refer-BDD decreases by ${-}1.78\%$, while the proposed EchoTrack increases by $1.44\%$, respectively, compared with those on the Refer-KITTI+ dataset.}
%</response-r19>

\subsubsection{Application of TransRMOT to AR-MOT} 
Table~\ref{lab:lab3} shows the results of applying the representative TransRMOT in RMOT to the AR-MOT task.
Audio integration into TransRMOT can be achieved through ASR or by directly employing HuBERT as a replacement for BERT to adapt it for AR-MOT. 
However, direct use of ASR cannot handle those interferences, increases costs (\#Params ${+}99.43$, FPS ${-}3.21$), and negatively impacts performance (HOTA ${-}3.51$, MOTA ${-}4.79$).
On the contrary, the proposed EchoTrack can maintain a high tracking capability while satisfying interaction convenience and efficiency in autonomous driving.

\begin{table}[]
\centering
\renewcommand\arraystretch{1.0}
\setlength\tabcolsep{7pt}
\caption{Ablation studies of using the TransRMOT~\cite{wu2023referring} as a baseline (BL) to check the effectiveness when the RMOT method is used for the AR-MOT.}
\resizebox{\linewidth}{!}{\begin{tabular}{l|ccccc}
\toprule [2pt]
Method      & HOTA$\uparrow$   & MOTA$\uparrow$  & IDF1$\uparrow$  & Param/M$\downarrow$  & FPS$\uparrow$  \\
\midrule [1pt]
BL + ASR    &  $\underline{33.63}$ & $8.62$  & $37.71$ & $449.36$ & $6.92$   \\
BL + HuBERT &   $33.58$ &   $\underline{8.69}$  & $\underline{39.06}$ &  $134.34$ &  $10.86$ \\

\midrule [1pt]
\rowcolor[gray]{.9} 
\textbf{EchoTrack} (ours)&     $\mathbf{37.14}$    &  $\mathbf{13.41}$    &  $\mathbf{44.30}$  & $143.79$ &  $10.13$     \\
\bottomrule [2pt]
\end{tabular}}
\label{lab:lab3}
\end{table}

\begin{table}[]
\centering
%<*response-r12-tab>
\setlength\tabcolsep{7pt}
\renewcommand\arraystretch{1.0}
\caption{\revised{Quantitative results of MOTRv2~\cite{zhang2023motrv2}, TransRMOT~\cite{wu2023referring}, and the proposed EchoTrack on the Echo-BDD under different environmental conditions.}}

\resizebox{\linewidth}{!}{\begin{tabular}{l|ccccc}
\toprule [2pt]
\multicolumn{1}{c|}{Methods} & \multicolumn{1}{c}{HOTA$\uparrow$}  & \multicolumn{1}{c}{DetA$\uparrow$}  & \multicolumn{1}{c}{AssA$\uparrow$}  & \multicolumn{1}{c}{MOTA$\uparrow$}  & \multicolumn{1}{c}{IDF1$\uparrow$}  \\

\toprule [1pt]
\multicolumn{6}{l}{\textbf{\revised{\textit{Rainy condition}}}}\\
MOTRv2~\cite{zhang2023motrv2}                   & $\underline{21.41}$ & $11.46$ & $\textbf{40.54}$ &  $\underline{-44.84}$ & $16.76$  \\
TransRMOT~\cite{wu2023referring}                   &  $20.01$  &  $\underline{12.64}$    &  $31.92$  &  $-47.65$  &    $\underline{17.70}$  \\
\rowcolor[gray]{.9}
\textbf{EchoTrack} (ours)          &  $\textbf{30.70}$       &      $\textbf{27.68}$         &  $\underline{34.27}$  &  $\textbf{5.94}$ &   $\textbf{34.01}$  \\
\toprule [1pt]
\multicolumn{6}{l}{\revised{\textbf{\textit{Daylight condition}}}} \\ 
MOTRv2~\cite{zhang2023motrv2} & $23.84$  & $10.70$ & $\underline{54.35}$ &  $-53.96$  &  $\underline{18.65}$  \\
TransRMOT~\cite{wu2023referring}    &   $\underline{26.31}$     &  $\underline{12.50}$   &  $\textbf{56.34}$  &      $\underline{-40.05}$   &   $-40.06$  \\
\rowcolor[gray]{.9} 
\textbf{EchoTrack} (ours)   & $\textbf{29.85}$ & $\textbf{19.71}$  &  $45.87$ & $\textbf{5.51}$ &  $\textbf{29.27}$  \\
\toprule [1pt]
\multicolumn{6}{l}{\revised{\textbf{\textit{Night condition}}}} \\
MOTRv2~\cite{zhang2023motrv2}                   & $\textbf{33.10}$ & $\underline{18.17}$ & $\textbf{61.01}$ & $\underline{-15.03}$ & $\underline{34.65}$ \\
TransRMOT~\cite{wu2023referring}                   &  $25.53$	& $16.28$ &	$40.65$  &   $-15.67$  &  $28.49$      \\
\rowcolor[gray]{.9}
\textbf{EchoTrack} (ours)    & $\underline{31.00}$  &      $\textbf{23.68}$        &       $\underline{41.43}$      &    $\textbf{0.58}$     & $\textbf{39.29}$    \\
\toprule [1pt]
\multicolumn{6}{l}{\revised{\textbf{\textit{Overexposure condition}}}} \\
MOTRv2~\cite{zhang2023motrv2}                   &   $\underline{19.40}$  & $\underline{10.53}$ & $\textbf{36.61}$ & $-99.30$ & $\underline{15.46}$ \\
TransRMOT~\cite{wu2023referring}               &  $15.68$	& $7.40$  &	$34.12$ &  $\underline{-47.78}$   & $11.27$  \\
\rowcolor[gray]{.9}
\textbf{EchoTrack} (ours)          &      $\textbf{22.83}$          &        $\textbf{15.09}$         &      $\underline{34.99}$      &   $\textbf{4.16}$   &   $\textbf{21.19}$    \\
\toprule [1pt]
\multicolumn{6}{l}{\revised{\textbf{\textit{Underexposure condition}}}} \\
MOTRv2~\cite{zhang2023motrv2}                   &   $33.86$   & $20.18$ & $\underline{57.21}$ & \underline{$-12.98$} & $27.80$ \\
TransRMOT~\cite{wu2023referring}   & $\underline{37.21}$	& \underline{$22.88$} & \textbf{$\mathbf{60.70}$} & $-38.85$ & \underline{$36.86$}                          \\
\rowcolor[gray]{.9}
\textbf{EchoTrack} (ours)  & $\textbf{41.61}$ &     $ \textbf{30.42}$         &   $57.16$      & $\textbf{6.73}$ & $\textbf{49.96}$ \\
\toprule [2pt]
\end{tabular}}

%</response-r12-tab>
\label{lab:sup7}
\end{table}

\begin{table}[t!]
\small
\centering
\renewcommand\arraystretch{0.75}
\setlength\tabcolsep{11pt}
\caption{Ablation studies of using the TransRMOT~\cite{wu2023referring} as a baseline (BL) to check the effectiveness when BL is used for the AR-MOT.}
\resizebox{\linewidth}{!}{\begin{tabular}{l|c|ccc}
\toprule [2pt]
Method & Mode     & HOTA$\uparrow$  & MOTA$\uparrow$ & IDF1$\uparrow$ \\
\midrule [0.5pt]
\multirow{2}{*}{Complexity}    & $5\geq$  & $35.36$  & $5.11$  &  $41.22$   \\
    & $5\,\,\textless$  & $39.21$  & $22.83$  &  $47.88$   \\
\midrule [0.5pt]
\multirow{2}{*}{Timbre} & Male & $36.45$  &  $9.56$ & $42.45$  \\
 & Female & $37.78$  & $16.95$  &  $46.00$ \\
\midrule [0.5pt]
\multirow{3}{*}{Speed}  
& $0.5 \times$ & $29.78$ & $-29.73$ &  $31.87$   \\
& $1.25 \times$  & $36.88$ & $7.94$   &  $42.70$ \\
& $1.5 \times$ & $35.46$ &  $2.80$  & $41.50$  \\
\midrule [0.5pt]
\multirow{2}{*}{Noise} 
& $1$db  & $35.15$ &  $6.42$  & $40.49$  \\
& $2$db  & $34.07$ &  $9.30$  &  $39.87$ \\
\midrule [0.5pt]       
TransRMOT~\cite{wu2023referring} & $4$db  & $29.89$ & $-15.19$ & $33.67$  \\
\rowcolor[gray]{.9}
\textbf{EchoTrack} (ours) & $4$db  & $\textbf{33.76}$ &  $\textbf{6.67}$  &  $\textbf{39.13}$ \\
\bottomrule [2pt]
\end{tabular}}
\label{sup:lab5}
\end{table}

\subsection{Ablation Study} \label{ablation}

\subsubsection{\revised{More Results under Different Environmental Conditions}} 
%<*response-r12>
Table~\ref{lab:sup7} summarizes the results of MOTRv2~\cite{zhang2023motrv2}, TransRMOT~\cite{wu2023referring} and EchoTrack for different ``scenes'', ``weather'', and ``illumination''. 
Two key observations can be drawn from the findings: \revised{1) The results of all methods under various conditions are significantly different, which indicates that the AR-MOT benchmarks increase the attributes of ``scene'', ``weather'', and ``illumination'' in real-life scenarios, which is necessary to improve the quality of the benchmark. 
2) The proposed EchoTrack demonstrates superior tracking performance and robustness across all environmental conditions (\textit{Daylight}, \textit{Rainy}, \textit{Night}, \textit{Overexposure}, and \textit{Underexposure}), which may benefit from the proposed Bi-FCFM and ACTL modules. However, it is noteworthy that all methods face considerable challenges in maintaining continuous tracking ability, particularly under heightened scene complexity.}
%</response-r12>

\subsubsection{Tracking Performance with Different Interference} 
%<*response-r17-1>
We also explore audio complexity, timbre, speed, and noise, revealing their impact on the tracking performance in Table~\ref{sup:lab5}. Thanks to Bi-FCFM and ACTL, EchoTrack maintains superior tracking even with higher interference, but TransRMOT does not. It indicates that only HuBERT fails to prevent audio interference, highlighting the vital role of high-quality audio for AR-MOT. \revised{These findings also highlight an intriguing observation: except for language complexity, the tracking performance of improved text-based RMOT methods is more sensitive to speed and noise compared to EchoTrack. This underscores the importance of the AR-MOT task in automated driving and its superiority over text-based RMOT methods.}
%</response-r17-1>

\begin{table}[t!]
\centering

%<*response-r23-tab1>
\renewcommand\arraystretch{1.0}
\setlength\tabcolsep{2pt}
\caption{\revised{Ablation studies of using the TransRMOT~\cite{wu2023referring} as a baseline (BL) with Bi-FCFM, ACTL, and both of them trained on Echo-KITTI. The Param indicates the number of parameters.}}
\resizebox{\linewidth}{!}{\begin{tabular}{l|cccccc}
\toprule [2pt]
Method      & HOTA$\uparrow$  & MOTA$\uparrow$ & IDF1$\uparrow$ & \revised{Param/M}$\downarrow$ & \revised{FLOPs/G}$\downarrow$ & \revised{FPS}$\uparrow$ \\
\midrule [1pt]
BL &  $33.58$   &  $8.69$  &   $39.06$   & \revised{$134.34$} & \revised{$430.71$}  & \revised{$10.77$} \\
$+$ Bi-FCFM &  $35.73$   &   $18.41$   &   $42.48$   & \revised{$142.61$} & \revised{$491.67$}  & \revised{$10.30$}  \\
$+$ ACTL  &  $34.06$     &  $12.42$  &   $39.47$     & \revised{$134.73$}  & \revised{$430.72$}  & \revised{$10.89$}  \\
\rowcolor[gray]{.9} 
$+$ Bi-FCFM + ACTL & $\mathbf{37.14}$     &  $\mathbf{13.41}$    &  $\mathbf{44.30}$    & \revised{$143.79$} & \revised{$491.67$}  & \revised{$10.13$}  \\
\bottomrule [2pt]
\end{tabular}}

%</response-r23-tab1>
\label{lab:lab4}
\end{table}

\begin{table}[t!]
\centering
%<*response-r23-tab2>
\setlength\tabcolsep{4pt}
\renewcommand\arraystretch{1.0}
\caption{Ablation studies of Bi-FCFM w/ or w/o the $\mathcal{G}_{a2v}$ and $\mathcal{G}_{v2a}$.}
\resizebox{\linewidth}{!}{\begin{tabular}{cc|ccccccc}
\toprule [2pt]
$\mathcal{G}_{a2v}$ & $\mathcal{G}_{v2a}$   & HOTA$\uparrow$ & MOTA$\uparrow$ & IDF1$\uparrow$ & Param/M$\downarrow$ & \revised{FLOPs/G}$\downarrow$ & FPS$\uparrow$ \\
\midrule [1pt]
& & $33.44$ & $11.10$  &   $37.67$ &    $142.22$  & \revised{$483.07$} &  $10.69$ \\
\Checkmark &  & $36.17$  &  $12.27$    &  $43.02$ &  $143.01$ & \revised{$489.46$} & $10.57$ \\
    & \Checkmark    & $36.26$  &  $8.40$  & $43.03$  &  $143.01$ & \revised{$485.28$} & $10.39$  \\
\rowcolor[gray]{.9} 
\Checkmark   & \Checkmark & $\mathbf{37.14}$  &  $\mathbf{13.41}$    &  $\mathbf{44.30}$ & $143.79$ & \revised{$491.67$} &  $10.13$ \\
\bottomrule [2pt]
\end{tabular}}
%</response-r23-tab2>
\label{lab:lab5}
\end{table}

\begin{table}[t!]
\centering
\renewcommand\arraystretch{1.0}
\setlength\tabcolsep{9pt}
\caption{The performance change by EchoTrack with ACTL loss ($\mathcal{L}_{act}$) under different weights on the Echo-KITTI dataset.}
\resizebox{\linewidth}{!}{\begin{tabular}{c|ccccc}
\toprule [2pt]
Weight      & HOTA$\uparrow$ & DetA$\uparrow$ & AssA$\uparrow$ & MOTA$\uparrow$ & IDF1$\uparrow$ \\
\midrule [1pt]
$0.5$    & $36.06$  & $26.18$  & $51.11$ &  $9.06$ &  $43.35$    \\
$1.0$    &  $36.35$  & $26.85$ & $50.56$ &  $12.33$  & $43.81$  \\
\rowcolor[gray]{.9}
$2.0$    & $\mathbf{37.14}$     &  $\mathbf{27.39}$    &  $\mathbf{51.88}$    &  $\mathbf{13.41}$    &  $\mathbf{44.3}$       \\
$4.0$    & $36.31$     &  $27.61$ &  $49.02$ &  $10.46$    & $43.32$    \\
\bottomrule [2pt]
\end{tabular}}
\label{lab:lab6}
\end{table}

\subsubsection{Evaluation of Bi-FCFM and ACTL} 
%<*response-r17-2>
As shown in Table~\ref{lab:lab4}, with the equipping of Bi-FCFM with ACTL, the localization and tracking capabilities of baseline are effectively improved. It brings the BL a gain of $3.56$, $4.72$, and $5.24$ on HOTA, MOTA, and IDF1, respectively.
In addition, to explicitly confirm the role of components in Bi-FCFM, we conducted the experiments shown in Table~\ref{lab:lab5}.
\revised{The statistical results show that Bi-FCFM greatly enhances the performance of baseline (\#HOTA +3.7, MOTA +2.31) with only a negligible increase in computational overhead (\#Param +1.57M). These performance gains are attributed to the ability of the proposed Bi-FCFM to capture both frequency and source domain cues unique to text and audio. This capability enhances the model's processing of contextual information, resulting in consistent performance improvements in both AR-MOT and RMOT tasks. Additionally, regardless of whether audio- or text-based methods are used, there is a general attenuation of the referring features during long-range propagation. This issue is mitigated by ACTL, which effectively transfers the referring features across distances to the model output. Consequently, the integration of Bi-FCFM with ACTL not only enables the model to better process contextual information but also ensures consistent performance enhancements in both tasks.}
%</response-r17-2>

\subsubsection{Hyperparameter Settings for ACTL}
In Table~\ref{lab:lab6}, we analyze the impact of varying ACTL weights on the performance of EchoTrack. It is evident that as the weight incrementally rises from $0.5$ to $2.0$, the performance of EchoTrack steadily improves, reaching an optimal level at a weight of $2.0$. However, when the weight is increased to $4.0$, a marginal decline in the model's performance is observed. Consequently, the optimal weight assigned to ACTL is determined to be $2.0$.

\section{Conclusion}
\label{sec:conclusion}
In this work, we introduce a novel referring scene understanding task, \textit{i.e.}, Auditory Referring Multi-Object Tracking (AR-MOT), and a new AR-MOT method, \textit{i.e.}, EchoTrack. 
Our EchoTrack consists of two components designed for AR-MOT, \textit{i.e.}, the Bidirectional Frequency Domain Cross-attention Fusion Module (Bi-FCFM) for adaptive alignment and fusion of audio-visual features, and the Audio-visual Contrastive Tracking Learning (ACTL) to alleviate the loss of referring features in long-range propagation. 
Moreover, we have established the first set of large-scale AR-MOT benchmarks for autonomous driving, including Echo-KITTI, Echo-KITTI+, and Echo-BDD. 
Extensive experiments show the effectiveness of our proposed EchoTrack solution, consistently delivering state-of-the-art performance across various conditions.

In the future, we plan to further enhance the tracking performance when facing severe object motion and occlusion scenarios.
%<*response-r23>
\revised{Consequently, how to determine the motion states of the objects and designing more lightweight networks to reduce the computational demands of AR-MOT modeling represents a promising research avenue within AR-MOT.}
%</response-r23>
In addition, we intend to explore the possibility of unleashing the potential large language models for AR-MOT.

\bibliographystyle{IEEEtran}
\bibliography{bib}

\end{document}